\documentclass[preprint, 12pt]{elsarticle}

\usepackage{amssymb}
\usepackage{amsmath}
\usepackage{amsthm}

\usepackage{lineno}

\usepackage{amsfonts}
\usepackage{relsize}
\usepackage{url}
\usepackage{amssymb}
\usepackage{graphicx}
\usepackage{circuitikz}
\usepackage{fancyhdr}
\usepackage{stmaryrd}
\SetSymbolFont{stmry}{bold}{U}{stmry}{m}{n}
\usepackage{upgreek}
\usepackage{epstopdf}
\usepackage{mathrsfs}
\usepackage{algorithmic}

\usepackage{enumitem}
\setlist[enumerate]{leftmargin=.5in}
\setlist[itemize]{leftmargin=.5in}

\usepackage{amsopn}

\newtheorem{remark}{Remark}[subsection]
\newtheorem{theorem}{Theorem}

\newcommand{\ml}{\boldsymbol{\Lambda}}
\newcommand{\ms}{\boldsymbol{\Sigma}}
\newcommand{\mr}[1]{\mathbf{R}_{\text{#1}}}

\newcommand{\vx}{\mathbf{x}}
\newcommand{\vy}{\mathbf{y}}

\newcommand{\vf}{\mathbf{f}}
\newcommand{\vxt}{\mathbf{x}(t)}
\newcommand{\vyt}{\mathbf{y}(t)}

\newcommand{\vm}[1]{\boldsymbol{\mu}_{\text{#1}}}

\newcommand{\dt}{\Delta t}
\newcommand{\vw}{\mathbf{W}}

\newcommand{\cF}{\mathcal{F}}
\newcommand{\pp}{\mathbb{P}}

\newcommand{\ee}[1]{\mathbb{E}\left[#1\right]}
\newcommand{\nf}{\normalfont{f}}
\newcommand{\ns}{\normalfont{s}}
\newcommand{\rmd}{\mathrm{d}}

\allowdisplaybreaks
\journal{Physica D: Nonlinear Phenomena}

\begin{document}

\begin{frontmatter}



\title{Bridging Prediction and Attribution: Identifying Forward and Backward Causal Influence Ranges Using Assimilative Causal Inference} 


\author[label1]{Marios Andreou} 
\ead{mandreou@math.wisc.edu}
\ead[url]{https://mariosandreou.short.gy/Homepage}

\author[label1]{Nan Chen\corref{corr}}
\ead{chennan@math.wisc.edu}
\ead[url]{https://people.math.wisc.edu/~nchen29}

\affiliation[label1]{organization={Department of Mathematics, University of Wisonsin--Madison},
            addressline={480 Lincoln Drive}, 
            city={Madison},
            postcode={53706}, 
            state={WI},
            country={USA}}

\cortext[corr]{Corresponding Author}

\begin{abstract}
Causal inference identifies cause-and-effect relationships between variables. While traditional approaches rely on data to reveal causal links, a recently developed method, assimilative causal inference (ACI), integrates observations with dynamical models. It utilizes Bayesian data assimilation to trace causes back from observed effects by quantifying the reduction in uncertainty. ACI advances the detection of instantaneous causal relationships and the intermittent reversal of causal roles over time. Beyond identifying causal connections, an equally important challenge is determining the associated causal influence range (CIR), indicating when causal influences emerged and for how long they persist. In this paper, ACI is employed to develop mathematically rigorous formulations of both forward and backward CIRs at each time. The forward CIR quantifies the temporal impact of a cause, while the backward CIR traces the onset of triggers for an observed effect, thus characterizing causal predictability and attribution of outcomes at each transient phase, respectively. Objective and robust metrics for both CIRs are introduced, eliminating the need for empirical thresholds. Computationally efficient approximation algorithms to compute CIRs are developed, which facilitate the use of closed-form expressions for a broad class of nonlinear dynamical systems. Numerical simulations demonstrate how this forward and backward CIR framework provides new possibilities for probing complex dynamical systems. It advances the study of bifurcation-driven and noise-induced tipping points in Earth systems, investigates the impact from resolving the interfering variables when determining the influence ranges, and elucidates atmospheric blocking mechanisms in the equatorial region. These results have direct implications for science, policy, and decision-making.
\end{abstract}

\begin{graphicalabstract}
\vspace*{-0.8cm}
\begin{figure}[!ht]
  \makebox[\textwidth][c]{\includegraphics[width=1.0\textwidth]{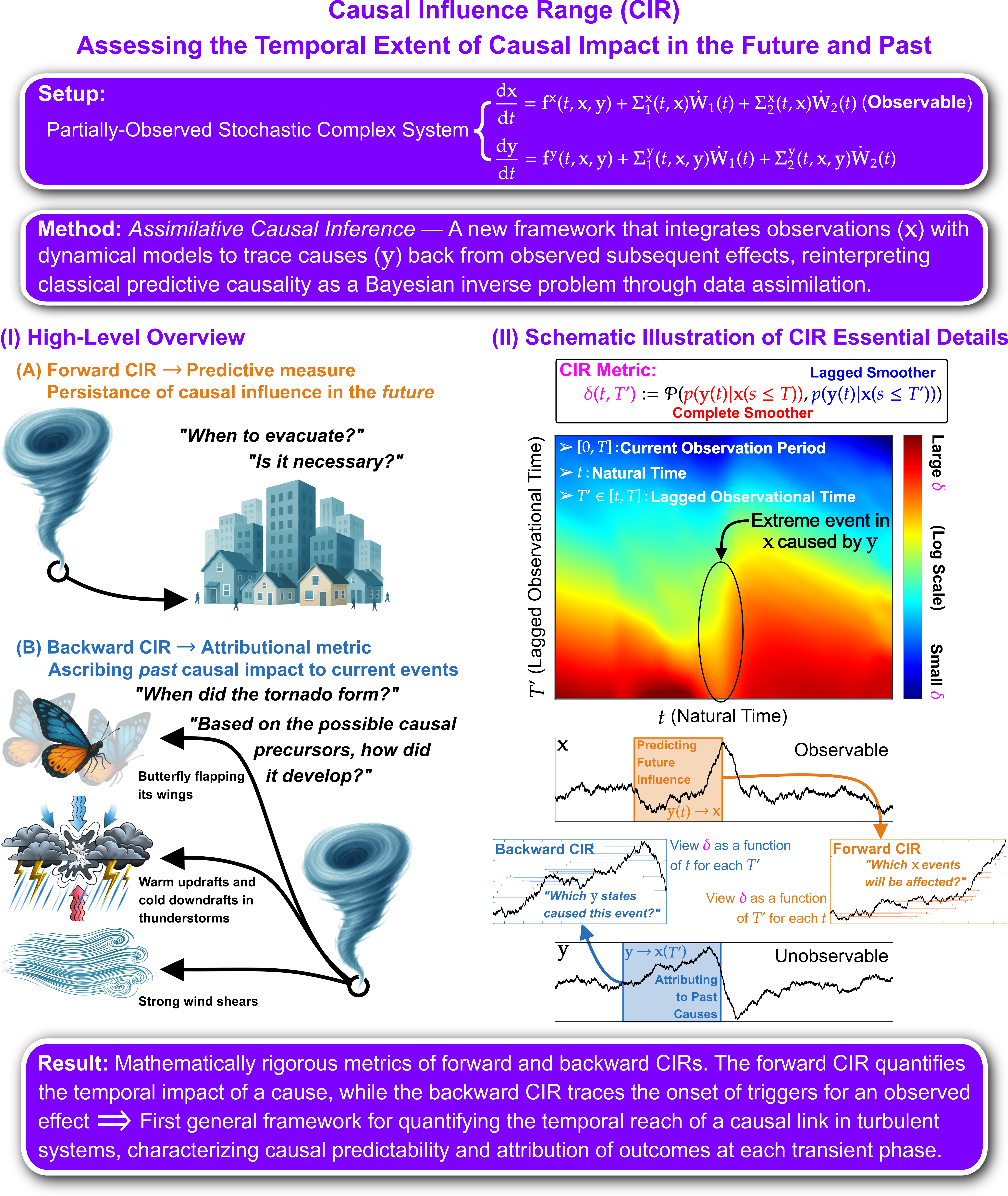}}%
\end{figure}
\end{graphicalabstract}

\begin{highlights}
\item Introduces a rigorous forward and backward causal influence range (CIR) theory
\item Builds upon assimilative causal inference, an inverse-problem viewpoint of causality
\item Develops threshold-free, robust, and efficient algorithms for computing both CIRs
\item Quantifies causal predictability and attribution in nonlinear dynamical systems
\item Applies the CIR framework to the analysis of tipping points and atmospheric blocking
\end{highlights}


\end{frontmatter}



\section{Introduction} \label{sec:Introduction}

Causal inference identifies cause-and-effect relationships between state variables \cite{pearl2009causality, morgan2014counterfactuals, scholkopf2017elements}. It is widely used across various scientific and societal areas. Beyond improving the understanding of the underlying dynamics, causal inference additionally plays a crucial role with assisting in the discovery of the governing equations, evaluating interventions, and informing decision-making \cite{runge2019detecting, charakopoulos2018dynamics, angrist2008mostly, shi2022learning, rubin1974estimating, rothman2005causation, weichwald2021causality, shams2022bayesian, zhang2011causality}.

Most of the traditional causal inference methods utilize multivariate time series data. Methods falling into this category include direct time series analysis (Granger causality \cite{granger1969investigating}, transfer entropy \cite{schreiber2000measuring, barnett2009granger}, causation entropy \cite{sun2014causation, sun2015causal}), state space reconstruction methods (topological causality \cite{harnack2017topological}, convergent cross mapping \cite{sugihara2012detecting, monster2017causal}) and statistical learning of structural causal models (interventional data methods \cite{shanmugam2015learning, brouillard2020differentiable}, mixture causal discovery \cite{varambally2023discovering}). In these approaches, the identified time-averaged causal relationships provide an overall indication of causality. However, many complex dynamical systems exhibit strong intermittency and nonlinearity, which means the cause-and-effect relationships between state variables may cease or shift repeatedly over time. Therefore, identifying instantaneous causal links is crucial for understanding transient behaviors such as extreme events, bifurcations, and regime switches. There exist model-based strategies that track time-dependent relationships, such as those utilizing functional operators (information flow \cite{liang2005information}, Koopman causality \cite{rupe2024causal}) and linear response theory (fluctuation-dissipation theorem \cite{falasca2024data}, Green's function \cite{lucarini2024detecting}). However, these methods are usually computationally intractable even when validating the hypothesis of a single possible causal scenario, while some of them are only able to provide temporal results for a very short time.

Assimilative causal inference (ACI) \cite{andreou2026assimilative}, an approach which aims to capture time-evolving causal relationships, has recently been developed. Using Bayesian data assimilation, it integrates data and models for causal inference, formulating the study of causality as an inverse problem of uncertainty quantification where causes are traced backward from observed effects. Key advantages of ACI include the online tracking of cause-and-effect roles which can intermittently swap, the accommodation of short and incomplete datasets that contain observations only from potential effects, and computational efficiency in high-dimensional settings by leveraging established techniques from data assimilation.

Although the discovery of instantaneous causal links is a critical first step, a comprehensive analysis must also quantify the associated causal influence range (CIR) that defines the temporal domain of significant causal impact. This concept has two distinct interpretations: A forward (predictive) CIR, describing the future period over which a current cause will remain influential, and a backward (attributable) CIR, describing the past period where causal progenitors responsible for a currently observed effect emerged.
\begin{enumerate}[label=\textbf{(\Alph*)}]
    \item The forward CIR quantifiably answers the operational question: “For how long will this causal effect persist?”. It assesses the remaining duration of an ongoing event or, following an early warning signal, infers the likelihood of an extreme event's onset. By providing a forecast of causal persistence, the forward CIR facilitates proactive preparedness for preventing catastrophic events, such as droughts, tsunamis, or cyclones.

    \item The backward CIR addresses the question: “When did the causal precursors for this observed event emerge?”. It pinpoints the onset of a damaging event and reveals its underlying driving mechanisms. By supporting root-cause analysis through causal attribution, the backward CIR guides targeted strategies for disaster preparedness and risk reduction. For instance, it can identify which factors most strongly influence long-term changes in Earth system processes, thereby helping to prioritize the most effective adaptation measures.
\end{enumerate}

This work builds upon the ACI framework by developing mathematically rigorous metrics for forward and backward CIRs. These formulations provide the first general framework for quantifying the temporal reach of a causal link in turbulent systems, both in the predictive (future) and attributable (past) directions. A key advantage of this framework is that it utilizes Bayesian data assimilation through ACI to yield computable expressions for CIRs. These expressions are applicable to complex dynamical systems, including those with diverse autocorrelation structures and chaotic dynamics. Critically, these CIR measures are objectively defined without relying on empirical thresholds. Furthermore, they admit computationally efficient algorithms with closed-form approximations for a broad class of turbulent systems, making asymptotic analysis and practical estimation possible even in high-dimensional applications.

The remainder of this paper is organized as follows. Section~\ref{sec:Methodology} provides a brief overview of ACI fundamentals and develops the theory for forward and backward CIRs, including derivations of computationally efficient approximations. Section~\ref{sec:Case_Studies} applies the CIR framework to: (a) Distinguishing bifurcation-driven from noise-induced tipping points in Earth system models, (b) clarifying prediction and attribution in multiscale atmospheric models by resolving interfering dynamical components, and (c) identifying atmospheric blocking mechanisms in an equatorial circulation model with flow-wave interactions and seasonal feedbacks. Concluding remarks and discussions are presented in Section~\ref{sec:Conclusion}. Additional mathematical derivations and technical details are given in the Appendix, where we also introduce a broad class of nonlinear dynamical systems that enables the analytical evaluation of CIR metrics without requiring ensemble approximations.

\section{Methodology} \label{sec:Methodology}

\subsection{Notations and Setup}
Throughout this work, boldface variables are used for multidimensional quantities. Lowercase variables denote column vectors, while uppercase ones denote matrices. The only exception is $\vw$ (with some subscript), which represents a vector-valued Wiener process due to literary tradition. Furthermore, for simplicity, we follow the notational convention from physics and do not distinguish between random variables and their realizations.

Let $\mathscr{B}=(\Omega, \cF, \mathbb{F}, \mathbb{P})$ be an augmented probability space for filtering on the time interval $[0,T]$, $T\in(0,+\infty)$. Denote by $\big(\mathbf{x}(t,\omega),\mathbf{y}(t,\omega)\big)\in\mathbb{R}^{k+l}$, for $t\in[0,T]$ and $\omega\in\Omega$, a $(k+l)$-dimensional partially-observable stochastic process on $\mathscr{B}$, where $\vx$ is the $k$-dimensional observable component while $\vy$ is the $l$-dimensional unobservable part. Without loss of generality, we consider real-valued processes, with $\vx$ being continuously observed (an analogous framework can be developed for discrete-time observations). For notational simplicity, we henceforth drop the sample space ($\Omega$) dependence, but it is always implied. We assume $(\vx,\vy)$ is adapted to the filtration $\mathbb{F}$ and that $\vx(s\leq t)$ represents a time series of $\vx$ over $[0,t]$, i.e., a realization for a fixed $\omega\in\Omega$. Finally, for explicitness, we hereafter write $\big(\boldsymbol{\cdot}\big|\vx(s\leq t)\big)$ to indicate the fact that we are conditioning on the $\sigma$-algebra generated by $\{\vx(s)\}_{s\leq t}$; nonetheless, throughout this work we assume that the observed time series of $\vx$ is free of observational noise, with noise-contaminated realizations (i.e., the effect of observational randomness on the conditional distributions) being left for consideration in future upcoming work.

\subsection{Fundamentals of Assimilative Causal Inference (ACI)} \label{sec:ACI}

We provide a brief review of the recently developed ACI framework \cite{andreou2026assimilative}. We assume that the evolution of $(\vx,\vy)$ is governed by \cite{rozovsky2012stochastic}:
\begin{subequations} \label{eq:general_CTNDS}
	\begin{align}
		\frac{\rmd\mathbf{x}(t)}{\rmd t} &= \mathbf{f}^{\vx}(t, \mathbf{x},\mathbf{y})+ \boldsymbol\Sigma^{\vx}_1(t,\vx)\dot{\mathbf{W}}_1(t)+\boldsymbol\Sigma^{\vx}_2(t,\vx)\dot{\mathbf{W}}_2(t), \label{eq:general_CTNDS1}\\
		\frac{\rmd\mathbf{y}(t)}{\rmd t} &= \mathbf{f}^{\vy}(t, \mathbf{x},\mathbf{y}) + \boldsymbol\Sigma^{\vy}_1(t,\vx,\vy)\dot{\mathbf{W}}_1(t)+\boldsymbol\Sigma^{\vy}_2(t,\vx,\vy)\dot{\mathbf{W}}_2(t), \quad t\in[0,T], \label{eq:general_CTNDS2}
	\end{align}
\end{subequations}
where $\dot{\vw}_1$ and $\dot{\mathbf{W}}_2$ are independent real-valued Gaussian white noises. The deterministic dynamics are allowed to be highly nonlinear in $(t,\vx,\vy)$, while uncertainty in the dynamical system is encoded through stochastic forcings that are possibly state-dependent (multiplicative) and cross-correlated. Certain regularity and identifiability conditions are enforced such that the conditional distribution of $\vy$ given the data $\vx$ contains all available information about $\vy$ over the considered observational time frame \cite{andreou2026assimilative, rozovsky2012stochastic}.

Bayesian data assimilation applies Bayes' theorem to combine a prior distribution for the current unobserved state $\vy(t)$, obtained by forward integrating \eqref{eq:general_CTNDS} (the forecast step), with the likelihood from a single realization of $\vx$. This yields a posterior distribution for $\vyt$ (the analysis step). Utilizing this, if we consider $\vx$ to be the subsequent effect of $\vyt$, then ACI identifies $\vyt$ as an instantaneous cause of $\vx$ at time $t\in[0,T]$ if the incorporation of future observations of $\vx$ from $(t,T]$ reduces the posterior uncertainty in estimating $\vyt$ beyond the reduction already achieved by using only its historical data, $\vx(s\leq t)$. This excess uncertainty reduction can be quantified using the relative entropy \cite{kleeman2011information, majda2018model}:
\begin{equation}  \label{eq:relative_entropy}
	\mathcal{P}\big(p_t^{\text{\normalfont{s}}}(\vy|\vx),p_t^{\text{\normalfont{f}}}(\vy|\vx)\big):=\int p_t^{\text{\normalfont{s}}}(\vy|\vx)\log\big(p_t^{\text{\normalfont{s}}}(\vy|\vx)/p_t^{\text{\normalfont{f}}}(\vy|\vx)\big)\rmd\vyt,
\end{equation}
where $p_t^{\text{\normalfont{s}}}(\vy|\vx):=p\big(\vyt|\vx(s\leq T)\big)$ and ${p_t^{\text{\normalfont{f}}}(\vy|\vx):=p\big(\vyt|\vx(s\leq t)\big)}$ are the smoother and filter posterior probability density functions (PDFs) of $\vyt$, respectively. This choice of statistical divergence is without loss of generality and chosen so purely on the basis of mathematical convenience \cite{andreou2026assimilative}. Therefore, whenever
\begin{equation}  \label{eq:RE_filter_smoother}
	\mathcal{P}\big(p_t^{\text{\normalfont{s}}}(\vy|\vx),p_t^{\text{\normalfont{f}}}(\vy|\vx)\big)>0,
\end{equation}
meaning there is information to be gained about $\vyt$'s state from incorporating the future data of $\vx$, we identify an assimilative causal relationship from $\vyt$ to $\vx$ \cite{andreou2026assimilative}, denoted as
\begin{equation} \label{eq:ACI_cause_notation}
	\vyt \rightarrow \vx.
\end{equation}

ACI differs from the classical notion of predictive causality as outlined by Granger \cite{granger1969investigating, granger1980testing}. ACI solves an inverse problem of uncertainty quantification in statistical inference. Under the Bayesian data assimilation pipeline, candidate causal variables are traced backward from data of their subsequent observed effects. As such, identifying cause-and-effect relationships is not based on extrapolating causes forward in time through \eqref{eq:general_CTNDS} to potentially obtain the resulting informational response on the target variables' entropy. ACI instead interpolates the effects onto their candidate causes in the past, bypassing the need for restrictive model structures, sufficiently long time series for temporal averages, or observation of the whole state space. The last of these is practically meaningful, as data from the causes to be identified may not always be available.

Expanding on this, \cite{andreou2026assimilative} develops an extension of the ACI framework that allows for the assessment of whether $\vyt$ causes a subset of the observed variables, $\vx_{\text{A}}$, where $\vx=(\vx_{\text{A}},\vx_{\text{B}})$, while in the presence of the remaining non-target variables, $\vx_{\text{B}}$. Here, $\vx_{\text{B}}$ can play the role of confounders, mediators, moderators, colliders, or instrumental variables in the underlying causal link (if it exists). This is sometimes referred to as a conditional causal relationship, where $\vyt$ instantaneously causes $\vx_{\text{A}}$ beyond the contributions of $\vx_{\text{B}}$. Technical details on conditional ACI are provided in Appendix~\ref{sec:Conditional_ACI}.

We note here that in this inverse-problem formulation of causality through ACI and in the forward and backward CIR theory developed in this work (see Section \ref{sec:CIRs}), we assume that the stochasticity solely stems from the random, nonlinear physical forcings in \eqref{eq:general_CTNDS} (due to the simplifying assumption of noise-free observed time series). Nevertheless, Bayesian data assimilation remains applicable whenever any nondegenerate uncertainty model is considered, even in conjunction with a deterministic, chaotic, partially-observed dynamical system, meaning the applicability of the proposed ACI and CIR frameworks is much broader than this stochastic setting. More generally, the assumed uncertainty can be epistemic, arising from incomplete knowledge through the initial conditions or model parameters and structure (e.g., unresolved processes) that is encoded in the prior, with positive Lyapunov exponents amplifying small initial-condition, discretization, and model errors along unstable directions, while the likelihood can account for aleatoric uncertainty due to partial, finite-resolution, or noisy observations \cite{law2015data,carrassi2022data}. Therefore, repeated Bayesian data assimilation is particularly important for uncertainty quantification and reduction, despite if the underlying dynamics are deterministic. As such, through careful treatment of the filter and smoother distributions of data assimilation, the same Bayesian-posterior-based ACI and CIR definitions (see Section \ref{sec:CIRs}) can be applied. Only in the ideal limit of a perfectly known model with complete, noiseless observations does Bayesian data assimilation become singular and so regularization for ACI and CIR analysis is necessary, since the posterior distributions can become degenerate.

\subsection{Causal Influence Range (CIR) Estimation Using ACI} \label{sec:CIRs}

In what follows, for the sake of simplicity, we discuss the theory of unconditional CIR through the lens of ACI, where a similar setup can be built for conditional CIR by substituting the appropriate equations (\eqref{eq:RE_filter_smoother}--\eqref{eq:ACI_cause_notation} in Section~\ref{sec:ACI}) with their conditional counterparts (\eqref{eq:ACI_cause_cond_notation}--\eqref{eq:filter_smoother_ancillary_inf_uncert} in Appendix~\ref{sec:Conditional_ACI}).

While the uncertainty reduction in the smoother solution relative to the filter one in \eqref{eq:RE_filter_smoother} identifies the causal relationship \eqref{eq:ACI_cause_notation}, it does not reveal the associated temporal extent of $\vy$'s causal impact on $\vx$. This time span is exactly the CIR of a causal relationship, as discussed in Section~\ref{sec:Introduction}. Depending on the adopted time directionality during the assessment of this temporal extent, there are two distinct formulations of CIR:
\begin{enumerate}[label=\textbf{(\Alph*)}]
\item Forward CIR (Section~\ref{sec:Forward_CIR}), which is the time window after $t$ that describes the future temporal extent of the causal state's, $\vyt$, influence on $\vx$. It defines the period for which $\vyt$ remains a significant predictor for the effect variables $\vx$. It characterizes the predictability and persistence of the causal variables.

\item Backward CIR (Section~\ref{sec:Backward_CIR}), which is the interval before $T$ that describes the past temporal extent of the currently observed effect's, $\vx(T)$, attribution to $\vy$'s impact. It identifies the period during which the causal variables', $\vy$, past states acted as $\vx(T)$'s precursors. It enables the attribution of an outcome to its historical causes, thus pinpointing its origin.
\end{enumerate}
Note that, in the context of the backward CIR, causal attribution proceeds backward from a fixed observational time, which is nevertheless arbitrary. We adopt the convention of denoting this reference time using $T>0$ for notational simplicity. Although the notation $\vx(t)$ offers a more direct analogy to the forward CIR, the use of $\vx(T)$ naturally facilitates the subsequent discussion of the backward CIR theory.

\subsubsection{A Unified ACI Framework for Assessing Forward and Backward CIRs} \label{sec:CIR_Framework}

In complex turbulent dynamical systems, memory effects (e.g., auto- and cross-correlations) necessarily decay over time. Consequently, in Bayesian data assimilation, new observations of $\vx$ typically influence the state estimation of $\vy$ only over a finite time window \cite{andreou2026adaptive}. Within the ACI framework, this fact implies that the causal variables $\vy$ effectively and measurably influence the states of the effect variables $\vx$ only for a limited duration. As a result, both the forward and backward CIRs are inherently finite in chaotic settings, enabling their rigorous mathematical treatment under a suitable formulation.

Let $T>0$ be the length of the total period of available observations, and let $t\in[0,T]$ be the current time. For every future time $T'\in[t,T]$, which we refer to as a lagged observational time (since it lags behind $T$), we define the following two posterior state estimations:
\begin{itemize}
	\item Complete smoother PDF, $p\big(\vyt\big|\vx(s\leq T)\big)$: The optimal state estimate of $\vyt$, conditioned on all available data up to time $T$.
	
	\item Lagged smoother PDF, $p\big(\vyt\big|\vx(s\leq T')\big)$: Suboptimal state estimation, conditioned only on the data up to time $T'$.
\end{itemize}
The statistical discrepancy between these PDFs can be intrinsically quantified through the relative entropy, without loss of generality, akin to \eqref{eq:RE_filter_smoother}:
\begin{equation} \label{eq:metric_CIR}
	0\leq \delta(t,T') := \mathcal{P}\big(p\big(\vyt\big|\vx(s\leq T)\big), p\big(\vyt\big|\vx(s\leq T')\big)\big), \quad 0\leq t \leq T'\leq T.
\end{equation}
We subsequently refer to $\delta(t,T')$ as the CIR metric, a function of both natural time $t$ and lagged observational time $T'$ (and of $T$, implicitly, but which we omit for notational simplicity). This quantity measures the lack of information in the state estimation of $\vyt$ when omitting the observational data of $\vx$ in $(T',T]$. This CIR metric is used to measure the actual CIRs. Specifically, under ACI's inverse-problem-based formulation of causality, the finite memory of the dynamics manifests itself in the temporal decay of:
\begin{enumerate}[label=\textbf{(\Alph*)}]
	\item $\vyt$'s influence on the future states of $\vx(T')$ as $T'$ increases from $t$ toward $T$. This is because increasing data incorporation leads to the CIR metric in \eqref{eq:metric_CIR} to decrease to zero pointwise due to properties of the relative entropy \cite{cai2002mathematical}. In this setting, the CIR metric in \eqref{eq:metric_CIR} is treated as a function of $T'$ with $t$ being fixed, and the region for which it is significant in $T'$ will define the forward CIR over $[t,T]$.
	
	\item $\vx(T)$'s attribution to the past states $\vy(t)$ as $t$ decreases from $T$, when $T'$ approaches $T$ from below. This is intuitively understood by discretizing $[0,T]$ with an observation rate of $\Delta{t}\ll1$, where $T'=T-\Delta{t}$ (see also Appendix~\ref{sec:Tool_Backward_CIR}). Here, the finite-memory effect similarly causes the CIR metric in \eqref{eq:metric_CIR} to diminish as $t$ decreases from $T$ during this limiting regime, with a causal relationship from $\vy$ to $\vx(T)$ being discovered if the additional or new observation of $\vx(T)$ provides significant uncertainty reduction in estimating the past states of $\vy$ beyond that already stemming from the data $\vx(s\leq T-\Delta t)$ as $\Delta t\rightarrow0^+$. In this backward-time context, the CIR metric in \eqref{eq:metric_CIR} is treated as a function of $t$, with arbitrary $T$ fixed and $T'$ sufficiently close to it such that the lagged smoother PDFs approximate the optimal complete smoother. The interval in natural time $t$ over which the CIR metric is significant as $T'\to T^-$ will define the backward CIR over $[0,T]$.
\end{enumerate}
Therefore, by appropriately treating the CIR metric in \eqref{eq:metric_CIR} as either a function of $T'$ or $t$ (for the latter case after the limit $T'\to T^-$), we can rigorously define ways to quantify the forward and backward CIR, respectively.

\begin{figure*}[!ht]%
\centering
\makebox[\textwidth][c]{\includegraphics[width=1.0\textwidth]{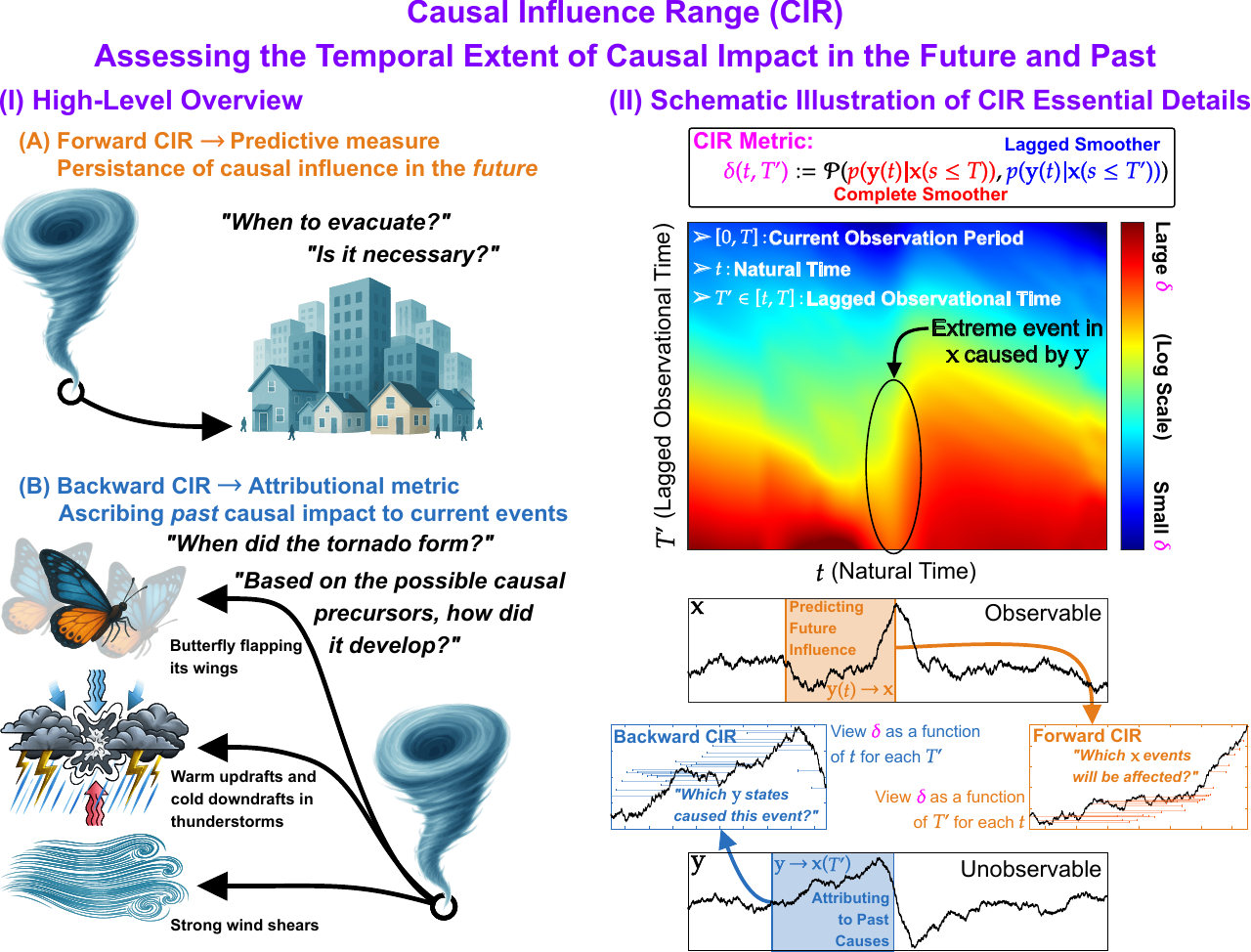}}
\caption{The ACI-based formulations of the forward and backward CIR measures. Panel (I): A high-level overview of the developed methodology via a real-world scenario (tornado forecast and attribution). Panel (II): Schematic illustration of the framework during an extreme event from a more technical viewpoint.}
\label{fig:schematic_diagram}
\end{figure*}

Panel (I) of Figure~\ref{fig:schematic_diagram} provides a high-level overview of the forward and backward CIRs, in the context of a real-world scenario. Panel (II) provides further details via a schematic diagram that outlines the essential mathematical aspects of the ACI-based CIR framework developed in this work during an extreme event.

\subsubsection{Inferring Causal Predictability: Forward CIR} \label{sec:Forward_CIR}

In this section, we develop a theoretical framework for quantifying the instantaneous forward CIR of an assimilative causal relationship ${\vyt\rightarrow\vx}$, at each time $t\in[0,T]$, thus assessing the future states of $\vx$ that $\vyt$ influences in $[t,T]$. Additional mathematical details and an illustration of the forward CIR mechanisms during the onset of an extreme event are provided in Appendix~\ref{sec:Forward_CIR_Details}, while physical and operational interpretations of the forward CIR are given in Appendix~\ref{sec:CIR_Connections}.

The forward CIR quantifies for how far into the future the causal state $\vyt$ influences $\vx$. To formalize this, we treat the CIR metric from \eqref{eq:metric_CIR} as a function of the lagged observational time $T'\in[t,T]$ for each $t\in[0,T]$. For analysis, we normalize its lagged observational time domain to $\mathrm{I}:=[0,1]$ via the transformation $T' = h(\tau) = t + \tau(T-t)$, defining the forward CIR metric:
\begin{equation} \label{eq:forward_CIR_metric}
	\delta^{\text{\normalfont{f}}}(\tau;t):=\delta(t, t+\tau(T-t)), \quad \tau\in \mathrm{I}.
\end{equation}
We denote its maximum information deficit at time $t$ by:
\begin{equation}  \label{eq:max_forward_CIR}
	M^{\text{\normalfont{f}}}(t):=\sup_{\tau\in\mathrm{I}}\big\{\delta^{\text{\normalfont{f}}}(\tau;t)\big\}>0, \quad t\in[0,T],
\end{equation}
which we assume exists and is nonzero, a condition satisfied in most practical settings. 

As alluded to in Section~\ref{sec:CIR_Framework}, since future observations of $\vx$ affect $\vyt$'s posterior state estimation only over a finite interval due to the finite-memory dynamics \cite{andreou2026adaptive}, $\delta^{\text{\normalfont{f}}}(\tau;t)$ generally decreases from its maximum, $M^{\text{\normalfont{f}}}(t)$, which in chaotic systems occurs at or shortly after $\tau=0$, down to zero at $\tau=1$; see Remark~\ref{rmk:forward_CIR_properties} and Figure~\ref{fig:details_FCIR}. This remark forms the basis for the forward CIR theory that follows.

The subjective forward CIR length of $\vyt\rightarrow\vx$, dependent on a threshold $\varepsilon$, is:
\begin{equation} \label{eq:subjective_forward_CIR_length}
    \overset{\sim}{\uptau}^{\,\text{\normalfont{f}}}(t,\varepsilon) := (T-t)\sup\big\{\tau\in\mathrm{I} : \delta^{\text{\normalfont{f}}}(\tau;t) > \varepsilon\big\}\in[0,T-t], \quad t\in[0,T],
\end{equation}
with $\overset{\sim}{\uptau}^{\,\text{\normalfont{f}}}(t,\varepsilon)=0$ if $\varepsilon\geq M^{\text{\normalfont{f}}}(t)$. The corresponding subjective forward CIR interval is:
\begin{equation} \label{eq:subjective_forward_CIR}
    \widetilde{\mathrm{CIR}}^{\,\text{\normalfont{f}}}(t,\varepsilon) := \big[t, t + \overset{\sim}{\uptau}^{\text{\,\normalfont{f}}}(t,\varepsilon)\big]\subseteq[t,T], \quad t\in[0,T].
\end{equation}

To remove the $\varepsilon$-threshold dependence from \eqref{eq:subjective_forward_CIR_length}--\eqref{eq:subjective_forward_CIR}, we define the objective forward CIR length by averaging over all $\varepsilon$:
\begin{equation} \label{eq:objective_forward_CIR_length}
    \uptau^{\,\text{\normalfont{f}}}(t) := \frac{1}{M^{\text{\normalfont{f}}}(t)}\int_0^{M^{\text{\normalfont{f}}}(t)} \overset{\sim}{\uptau}^{\,\text{\normalfont{f}}}(t,\varepsilon) \rmd\varepsilon, \quad t\in[0,T],
\end{equation}
with the corresponding objective forward CIR interval given by:
\begin{equation} \label{eq:objective_forward_CIR}
    \mathrm{CIR}^{\,\text{\normalfont{f}}}(t) := \big[t, t + \uptau^{\,\text{\normalfont{f}}}(t)\big]\subseteq[t,T], \quad t\in[0,T].
\end{equation}
The objective forward CIR interval always lies within the maximal subjective interval, while always containing the minimal one.

While there are multiple ways to resolve the $\varepsilon$-dependence of the subjective forward CIR length, which is the natural measure to predict the future extent of causality (see Section \ref{sec:CIR_Framework} and Appendix \ref{sec:Forward_CIR_Details} for the necessary arguments), the $\varepsilon$-averaging approach in \eqref{eq:objective_forward_CIR_length} is particularly appealing as it provides a simple and interpretable expression: by assigning a uniform weight to each subjective forward CIR length during integration, $1/M^{\text{\normalfont{f}}}(t)$, the essential temporal causal influence structure of $\vyt\rightarrow\vx$ is captured, thus the subjectivity in the intuitive or natural definition of the threshold-dependent forward CIR is impartially resolved, hence the objective designation. Operationally, this objective definition is corroborated by the fact that it: (a) is dimensionally consistent (time units), (b) guarantees consistent values within the range $[0,T-t]$, and (c) is equivalent to a computationally efficient time integral of the normalized forward CIR metric under suitable conditions often satisfied in practice for complex turbulent systems (see Section~\ref{sec:Forw_CIR_Approx}).

\subsubsection{Efficient Computation of the Objective Forward CIR} \label{sec:Forw_CIR_Approx}

Computationally, we can efficiently (lower-bound) approximate the objective forward CIR length in \eqref{eq:objective_forward_CIR_length} without quadrature over $\varepsilon$, through the following integral of the CIR metric in \eqref{eq:metric_CIR} over the lagged observational time $T'$ \cite{andreou2026assimilative}:
\begin{equation} \label{eq:forward_CIR_approximation}
    \uptau^{\,\text{\normalfont{f}}}_{\text{approx}}(t):= \frac{T-t}{M^{\text{\normalfont{f}}}(t)}\int_0^1 \delta^{\text{\normalfont{f}}}(\tau;t)\rmd \tau=\frac{1}{M^{\text{\normalfont{f}}}(t)}\int_t^T \delta(t,T')\rmd T' \leq \uptau^{\,\text{\normalfont{f}}}(t),
\end{equation}
with equality in \eqref{eq:forward_CIR_approximation} if and only if $\delta^{\text{\normalfont{f}}}(\boldsymbol{\cdot};t)$ is nonincreasing on $\mathrm{I}$. In this case, its maximum information deficit occurs at $\tau=0$:
\begin{equation*}
    {M^{\text{\normalfont{f}}}(t)=\delta^{\text{\normalfont{f}}}(0;t)}=\delta(t,t)=\mathcal{P}(p_t^{\text{\normalfont{s}}}(\vy|\vx),p_t^{\text{\normalfont{f}}}(\vy|\vx)),
\end{equation*}
which equals the ACI metric assessing $\vyt\rightarrow\vx$ in \eqref{eq:RE_filter_smoother}. Dimensional analysis confirms $\uptau^{\,\text{\normalfont{f}}}_{\text{approx}}(t)$ maintains consistent time units, ensuring the physical validity of this approximation.

\begin{remark} \label{rmk:approx_obj_forw_CIR_interval}
Since \eqref{eq:forward_CIR_approximation} is an approximate underestimate, it yields shorter objective forward CIR intervals compared to the exact definition.
\end{remark}

\subsubsection{Identifying Causal Attribution: Backward CIR} \label{sec:Backward_CIR}

In an analogous manner, in this section, we formulate the theoretical basis for quantifying the instantaneous backward CIR of an assimilative causal relationship of the form $\vy\rightarrow\vx(T)$, at an arbitrary observational time ${T>0}$, thus inferring the past states of $\vy$ in $[0,T]$ that act as causal precursors to the observed effect $\vx(T)$. Based on \eqref{eq:backward_CIR_metric}--\eqref{eq:max_backward_CIR} and the note concerning \eqref{eq:max_backward_CIR_monotonicity} in Theorem~\ref{thm:obj_backward_CIR_approx}, $\vy\rightarrow\vx(T)$ can be considered as a valid assimilative causal relationship when evidence for $\vy(T')\rightarrow\vx$ as $T'\to T^-$ are strong (at least above the baseline value of the initial-time information deficit, i.e., value of the CIR metric at $t=0$; see \eqref{eq:max_backward_CIR_monotonicity}). Further technical details and an illustration of the backward CIR mechanisms at the peak of an extreme event are presented in Appendix~\ref{sec:Backward_CIR_Details}. Physical and operational interpretations of the backward CIR are given in Appendix~\ref{sec:CIR_Connections}, alongside a summary of the mathematical relationship between the forward and backward CIR theory counterparts.

Whereas forward CIR measures future influence, backward CIR attributes observed effects at time $T$, $\vx(T)$, to past causes $\vy$. To quantify this, we treat the CIR metric from \eqref{eq:metric_CIR} as a function of natural time $t\in[0,T']$ for each $T'\in[0,T]$; as noted before, $T$ is fixed but arbitrary. For analysis, we normalize its natural time domain to $\mathrm{I}=[0,1]$ via the transformation ${t = g(\tau) = \tau T'}$ and then center it relative to its baseline value at $t=0$, defining the backward CIR metric:
\begin{equation} \label{eq:backward_CIR_metric}
    \delta^{\text{\normalfont{b}}}(\tau;T'):=|\delta(\tau T', T')-\delta(0,T')|, \quad \tau\in \mathrm{I}.
\end{equation}
The absolute value ensures nonnegativity, although $\delta(t,T')-\delta(0,T')$ is typically positive for $t$ sufficiently far from zero when $T'$ is close to $T$. As described in point \textbf{(B)} of Section~\ref{sec:CIR_Framework}, we focus on the limiting regime $T'\to T^-$, or limit of complete information incorporation, assuming the complete backward CIR metric
\begin{equation} \label{eq:complete_backward_CIR_metric}
    \lim_{T'\to T^-}\delta^{\text{\normalfont{b}}}(\tau;T'),
\end{equation}
exists and is well-defined for each $\tau\in\mathrm{I}$, with its maximum information deficit at time $T$ under $T'\to T^-$ denoted by:
\begin{equation} \label{eq:max_backward_CIR}
    M^{\text{\normalfont{b}}}(T):=\sup_{\tau\in\mathrm{I}}\Big\{\lim_{T'\to T^-}\delta^{\text{\normalfont{b}}}(\tau;T')\Big\}>0, \quad T>0,
\end{equation}
which we likewise assume that it exists and is nonzero, both valid assumptions in most applications.

\begin{remark} \label{rmk:continuity_back_CIR_metric}
Although $\delta^{\text{\normalfont{b}}}(\tau;T)=0$ by relative entropy properties \cite{cai2002mathematical}, $\lim_{T'\to T^-}\delta^{\text{\normalfont{b}}}(\tau;T')$ generally differs from zero. That is, the backward CIR metric is not necessarily left-continuous at $T$ as a function of $T'$. This discontinuity becomes pronounced when approximating the limit with discrete-time observations at small observation rates which is what we do for the case studies in Section~\ref{sec:Case_Studies}, with computational details given in Appendix~\ref{sec:Tool_Backward_CIR}.
\end{remark}

Recall from Section~\ref{sec:CIR_Framework}, that the uncertainty reduction in estimating $\vyt$ from $\vx(s\leq T)$ beyond $\vx(s\leq T-\Delta{t})$, with $\Delta t\ll 1$ a discrete time step, diminishes as $t$ decreases from $T$ due to the finite-memory dynamics \cite{andreou2026adaptive}. This indicates that $\lim_{T'\to T^-}\delta^{\text{\normalfont{b}}}(\tau;T')$ precisely measures the role of the effect $\vx(T)$ in state estimation, while under ACI's viewpoint it quantifiably traces it back to candidate causes $\vyt$ earlier in time $t\leq T$. As such, it generally decreases from its maximum, $M^{\text{\normalfont{b}}}(T)$, which in chaotic systems occurs at or shortly before $\tau=1$, down to zero at $\tau=0$; see Remark~\ref{rmk:backward_CIR_properties} and Figure~\ref{fig:details_BCIR}. Utilizing this fact, the backward CIR theory can be developed.

The subjective backward CIR length of $\vy\to \vx(T)$, dependent on a threshold $\varepsilon$, is:
\begin{equation} \label{eq:subjective_backward_CIR_length}
    \overset{\sim}{\uptau}^{\,\text{\normalfont{b}}}(T,\varepsilon) := T\Big(1-\sup\Big\{\tau\in\mathrm{I} : \lim_{T'\to T^-}\delta^{\text{\normalfont{b}}}(\tau;T') \leq \varepsilon\Big\}\Big)\in[0,T], \quad T>0,
\end{equation}
with $\overset{\sim}{\uptau}^{\,\text{\normalfont{b}}}(T,\varepsilon)=0$ when $\varepsilon< 0$. The corresponding subjective backward CIR interval is:
\begin{equation} \label{eq:subjective_backward_CIR}
    \widetilde{\mathrm{CIR}}^{\,\text{\normalfont{b}}}(T,\varepsilon) := \big[T-\overset{\sim}{\uptau}^{\,\text{\normalfont{b}}}(T,\varepsilon), T\big]\subseteq[0,T], \quad T>0.
\end{equation}

Similar to \eqref{eq:objective_forward_CIR_length}, averaging over all possible $\varepsilon$ thresholds gives the objective backward CIR length:
\begin{equation} \label{eq:objective_backward_CIR_length}
    \uptau^{\,\text{\normalfont{b}}}(T) := \frac{1}{M^{\text{\normalfont{b}}}(T)}\int_0^{M^{\text{\normalfont{b}}}(T)} \overset{\sim}{\uptau}^{\,\text{\normalfont{b}}}(T,\varepsilon) \rmd\varepsilon, \quad T>0,
\end{equation}
with the corresponding objective backward CIR interval defined as:
\begin{equation} \label{eq:objective_backward_CIR}
    \mathrm{CIR}^{\,\text{\normalfont{b}}}(T) := \big[T- \uptau^{\,\text{\normalfont{b}}}(T),T\big]\subseteq[0,T], \quad T>0.
\end{equation}
As with the forward CIR, the objective backward CIR interval always lies within the maximal subjective interval and contains the minimal one.

The rationale behind the objective backward CIR definition in \eqref{eq:objective_backward_CIR_length}--\eqref{eq:objective_backward_CIR} is analogous to that of the forward CIR in \eqref{eq:objective_forward_CIR_length}--\eqref{eq:objective_forward_CIR}, as described in detail at the end of Section~\ref{sec:Forward_CIR}. By averaging over all possible subjective backward CIR lengths, which inherently attribute the past interval of causality (see Section \ref{sec:CIR_Framework} and Appendix \ref{sec:Backward_CIR_Details} for the necessary arguments), the essential temporal causal influence structure of $\vy\rightarrow\vx(T)$ is captured. Likewise, this objective backward CIR definition enjoys dimensional consistency, consistent values in $[0,T]$, and equivalence to a computationally efficient time integral of the normalized complete backward CIR metric under suitable and similar monotonicity conditions (see Section~\ref{sec:Backward_CIR_Computation} and Theorem \ref{thm:obj_backward_CIR_approx}).

\subsubsection{Efficient Computation of the Objective Backward CIR} \label{sec:Backward_CIR_Computation}

Direct computation of the objective backward CIR length in \eqref{eq:objective_backward_CIR_length} is often prohibitive. Its subjective counterpart in \eqref{eq:subjective_backward_CIR_length} rarely admits analytical solutions, requires repeated smoother computations with each new observation of $\vx$, and involves quadrature that scales at best quadratically with the number of discretization points.

Mirroring \eqref{eq:forward_CIR_approximation}, we develop a computationally efficient (upper-bound) approximation to \eqref{eq:objective_backward_CIR_length} that avoids $\varepsilon$-quadrature and becomes exact when the complete backward CIR metric is nondecreasing on $\mathrm{I}$, which is often true in practice. The theorem below requires only measurability of $\lim_{T'\to T^-}\delta^{\text{\normalfont{b}}}(\boldsymbol{\cdot};T')$ over $\mathrm{I}$ (see Remark~\ref{rmk:back_CIR_measurability}), with its proof given in Appendix~\ref{sec:Comp_Back_CIR_Thm_Proof}.

\begin{theorem}[Computationally Efficient Approximation of the Objective Backward CIR]
    \label{thm:obj_backward_CIR_approx}
    Assume ${M^{\text{\normalfont{b}}}(T)=\sup_{\tau\in\mathrm{I}}\{\lim_{T'\to T^-}\delta^{\text{\normalfont{b}}}(\tau;T')\}>0}$ exists for each $T>0$. Then:
    \begin{equation}
        \begin{aligned}
            \uptau^{\,\text{\normalfont{b}}}_{\text{\normalfont{approx}}}(T)&:=\frac{T}{M^{\text{\normalfont{b}}}(T)}\int^1_0\lim_{T'\to T^-}\delta^{\text{\normalfont{b}}}(\tau;T')\rmd \tau=\frac{1}{M^{\text{\normalfont{b}}}(T)}\int_0^T \lim_{T'\to T^-} |\delta(t, T')-\delta(0,T')|\rmd t\\
            &=\frac{T}{M^{\text{\normalfont{b}}}(T)}\int_0^{M^{\text{\normalfont{b}}}(T)}\Big(1-\lambda_{\mathrm{I}}\Big(\Big\{\tau\in\mathrm{I} : \lim_{T'\to T^-}\delta^{\text{\normalfont{b}}}(\tau;T') \leq \varepsilon\Big\}\Big)\Big) \rmd\varepsilon
            \geq \uptau^{\,\text{\normalfont{b}}}(T),
        \end{aligned} \label{eq:obj_backward_CIR_approx}
    \end{equation}
    where $\lambda_{\mathrm{I}}$ denotes the Lebesgue measure on $\mathrm{I}$, with equality in \eqref{eq:obj_backward_CIR_approx} if and only if $\lim_{T'\to T^-}\delta^{\text{\normalfont{b}}}(\boldsymbol{\cdot};T')$ is nondecreasing on $\mathrm{I}$. In this case, its maximum information deficit occurs at $\tau=1$:
    \begin{equation}
        \begin{aligned}
        M^{\text{\normalfont{b}}}(T)&=\lim_{T'\to T^-}|\delta(T',T')-\delta(0,T')|\\
        &=\lim_{T'\to T^-}\big|\mathcal{P}\big(p_{T'}^{\text{\normalfont{s}}}(\vy|\vx),p_{T'}^{\text{\normalfont{f}}}(\vy|\vx)\big)-\mathcal{P}\big(p(\vy(0)|\vx(s\leq T)), p(\vy(0)|\vx(s\leq T'))\big)\big|,
    \end{aligned} \label{eq:max_backward_CIR_monotonicity}
    \end{equation}
    where the first term inside the absolute value quantifies the causal intensity of $\vy(T')\rightarrow \vx$, while the second measures the initial-time information deficit from limited data assimilation (due to omitting the $\vx$ data in $(T',T]$).
\end{theorem}

Thus, we can (upper-bound) approximate the $\varepsilon$-average in \eqref{eq:objective_backward_CIR_length} with a computationally cheaper time integral of the complete backward CIR metric instead, which becomes exact under certain monotonicity conditions. Furthermore, note that dimensional analysis confirms $\uptau^{\,\text{\normalfont{b}}}_{\text{approx}}(T)$ maintains consistent time units, ensuring its physical validity.

\begin{remark} \label{rmk:approx_obj_back_CIR_interval}
Since \eqref{eq:obj_backward_CIR_approx} is an approximate overestimate, it yields longer objective backward CIR intervals compared to the exact definition.
\end{remark}

\subsubsection{Asymptotic Temporal Behavior of the Objective Backward CIR in Conditionally Linear Systems} \label{sec:Backward_CIR_Linear}

Analytical study of the forward CIR metric $\delta^{\text{\normalfont{f}}}(\tau;t)$ in \eqref{eq:forward_CIR_metric} is challenging, since the CIR metric in \eqref{eq:metric_CIR} can exhibit erratic behavior when viewed as a function of the lagged observational time $T'$. This reflects the inherent difficulty of extrapolating temporal causal impact into the future. In contrast, the backward CIR metric $\delta^{\text{\normalfont{b}}}(\tau;T')$ in \eqref{eq:backward_CIR_metric} and its complete variant in \eqref{eq:complete_backward_CIR_metric} are more analytically tractable. By focusing on causal attribution rather than prediction and by leveraging the efficient approximation in Theorem~\ref{thm:obj_backward_CIR_approx}, we can establish rigorous asymptotic results ($T>0$ sufficiently large) for specific system classes.

We consider a conditionally linear system (i.e., linear in the unobserved variable) with state-independent feedbacks:
\begin{align}
    \begin{split}
        \rmd x &= (\lambda^xy+f^x(t,x))\rmd t+\sigma^x\rmd W_1,\\
        \rmd y &= (\lambda^yy+f^y(t,x))\rmd t+\sigma^y\rmd W_2,
    \end{split} \label{eq:reduced_model}
\end{align}
where $x$ is observed while $y$ is unobserved. Despite the conditionally linear structure of \eqref{eq:reduced_model}, the system remains highly nonlinear in $x$, being able to capture many complex dynamical regimes (see Appendix~\ref{sec:Linear_Dynamics_y} for concrete examples). The proof of the following theorem is given in Appendix~\ref{sec:Back_CIR_Linear_Thm_Proof}.

\begin{theorem}[Asymptotic Temporal Behavior of the Objective Backward CIR Length for Conditionally Linear Systems]
    \label{thm:obj_backward_CIR_linear}
    Consider \eqref{eq:reduced_model} under the assumptions of Theorem~\ref{thm:obj_backward_CIR_approx}, including the existence of filter and smoother equilibrium distributions. For large $T$ (which need not be that large in practice due to the conditional linearity), the approximate objective backward CIR satisfies:
    \begin{equation} \label{eq:obj_backward_CIR_linear}
        \uptau^{\,\text{\normalfont{b}}}_{\text{approx}}(T)=\Theta(1)\, \Leftrightarrow\, A_1 \leq \uptau^{\,\text{\normalfont{b}}}_{\text{approx}}(T)\leq A_2,
    \end{equation}
    where $A_1$ and $A_2$ are $T$-independent constants.
\end{theorem}

\begin{remark}
    Whenever Theorem~\ref{thm:obj_backward_CIR_linear} applies in practice, $\uptau^{\,\text{\normalfont{b}}}_{\text{approx}}(T)$ oscillates intermittently between the $T$-constant bounds in \eqref{eq:obj_backward_CIR_linear} due to the inherent dynamical noise in \eqref{eq:reduced_model}. See Figures~\ref{fig:climate_tip_fig_2}(c) and~\ref{fig:climate_tip_fig_4}(c) in Section~\ref{sec:Climate_Tipping} for a numerical confirmation of this behavior.
\end{remark}

\section{CIRs for Probing Complex Dynamical Systems} \label{sec:Case_Studies}

In this section, ACI and its CIR formulations are utilized to address three key scientific challenges: (a) Understanding bifurcation-driven and noise-induced tipping points in a conceptual Earth system model, (b) resolving interfering observed effects to improve prediction and attribution in low-frequency atmospheric variability models, and (c) investigating atmospheric blocking and unblocking mechanisms using a conceptual equatorial circulation model with flow-wave interactions and seasonal feedbacks. While the (conditional) ACI framework in \cite{andreou2026assimilative} and its forward and backward CIR theory that has been developed in Section \ref{sec:CIRs} assume general nonlinear stochastic dynamical systems, see \eqref{eq:general_CTNDS}, the models we study here belong in the conditional Gaussian nonlinear system (CGNS) family, which is a computational tool with highly nonlinear and complex dynamics that enjoys closed-form expressions for the (conditional) ACI and CIR metrics. Therefore, errors stemming from ensemble- or particle-based approximations of the posterior distributions are ignored for an unbiased analysis \cite{jiang2026continuous}. Implementation details are given in Appendix~\ref{sec:Tool_CGNS}. The conditional ACI framework outlined in Appendix~\ref{sec:Conditional_ACI} and the corresponding formulation of conditional CIRs are employed when studying conditional causal relationships. Finally, all objective forward and backward CIRs in the following figures are computed using the computationally efficient approximations in \eqref{eq:forward_CIR_approximation} and \eqref{eq:obj_backward_CIR_approx}, respectively.

\subsection{Studying Bifurcation-Driven and Noise-Induced Tipping Points in a Model with Intermittent Instabilities} \label{sec:Climate_Tipping}

Tipping points in Earth system models explain dynamical transitions between physical regimes, but distinguishing whether these are driven by external forcings (e.g., Milankovitch cycles \cite{berger1988milankovitch}) or internal variability remains a challenging task.

Mathematically, tipping points are analyzed through bifurcation structures where quasi-static attractors lose stability, potentially leading to new attractors, transient excursions, or equilibrium departure \cite{thompson2011predicting}. Understanding these mechanisms is crucial for assessing responses in Earth processes under perturbations \cite{abramov2009new}. We focus on two types of tipping points \cite{ashwin2012tipping}. First, on bifurcation-driven tipping, which occurs when control parameters cross critical thresholds due to smooth perturbations \cite{lenton2008tipping}. Early warning signals like rising autocorrelation and variance often precede tipping but cannot reliably distinguish genuine precursors from internal variability artifacts \cite{brovkin2021past}. Furthermore, in multiscale systems, bifurcations in unobserved fast components may escape detection, where rapid exogenous changes can outpace early warning signals in inertial systems \cite{kuehn2011mathematical}. ACI-based CIR metrics overcome these limitations: Forward CIR enables real-time inference of impending tipping without solely relying on observations or low-order statistical moments, while backward CIR supports post-tipping attribution to triggering precursors. Second, we analyze noise-induced tipping, which is driven by random fluctuations or internal variability. These transitions lack clear early warning signals since noise can trigger regime shifts without potential changes. In Earth models, explained via stochastic dynamical systems \cite{ghil2012topics}, weather modes can act as stochastic perturbations that excite atmospheric processes through stochastic resonance \cite{gammaitoni1998stochastic}. Forward and backward CIR measures can jointly enable efficient detection, anticipation, and attribution of noise-induced regime transitions.

These tipping points are studied using the following system:
\begin{subequations} \label{eq:climate_model}
    \begin{align}
        \frac{\mathrm{d}x}{\rmd t}&=x-d_xx^3-\alpha y+\sigma_x\dot{W}_x, \label{eq:climate_model1}\\
        \frac{\mathrm{d}y}{\rmd t}&=\gamma x+\beta-\frac{d_y}{\varepsilon}y+\frac{\sigma_y}{\sqrt{\varepsilon}}\dot{W}_y, \label{eq:climate_model2}\\
        \frac{\mathrm{d}\gamma}{\rmd t}&=-d_{\gamma}(\gamma-\bar{\gamma})+\sigma_{\gamma}\dot{W}_{\gamma}. \label{eq:climate_model3}
    \end{align}
\end{subequations}
This model can mimic essential structural features of low-frequency variability present in general circulation models not captured by linear Gaussian approximations~\cite{majda2010low}. The small parameter $0<\varepsilon<1$ separates the fast weather modes ($y$) from the slow ($x$) timescales, while $\gamma$ provides stochastic parameterization with multiplicative noise. We apply ACI and CIR to study:
\begin{itemize}
    \item {Bifurcation-driven tipping}:  In \eqref{eq:climate_model1}, $y$ acts as a random forcing on $x$, intermittently triggering instability or regime shifts by pushing the system beyond its global stability threshold. Forward CIR analysis of $y\rightarrow x|\gamma$ assesses future tipping risk, while backward CIR attributes tips to $y$'s past behavior as a bifurcation parameter.
    \item {Noise-induced tipping}: The colored noise introduced by $\gamma$ in \eqref{eq:climate_model3} reduces system memory relative to additive-noise alternatives, due to the turbulent dynamics generated \cite{lindner2004effects, casado1997noise}. Simultaneous study of the forward and backward CIRs for $\gamma\rightarrow y|x$ helps identify these difficult-to-detect regime shifts linked to low-frequency variability.
\end{itemize}
The following parameter values are used in this study: $\varepsilon\in\{0.01,0.1\}$, ${d_x=1/3}$, $\alpha=4$, $\sigma_x=0.2$, $d_y=0.2$, $\beta=-0.8$, $\sigma_y=0.3$, $d_{\gamma}=0.5$, $\bar{\gamma}=1$, $\sigma_{\gamma}=2$.

\begin{figure*}[!ht]%
\centering
\includegraphics[width=\textwidth]{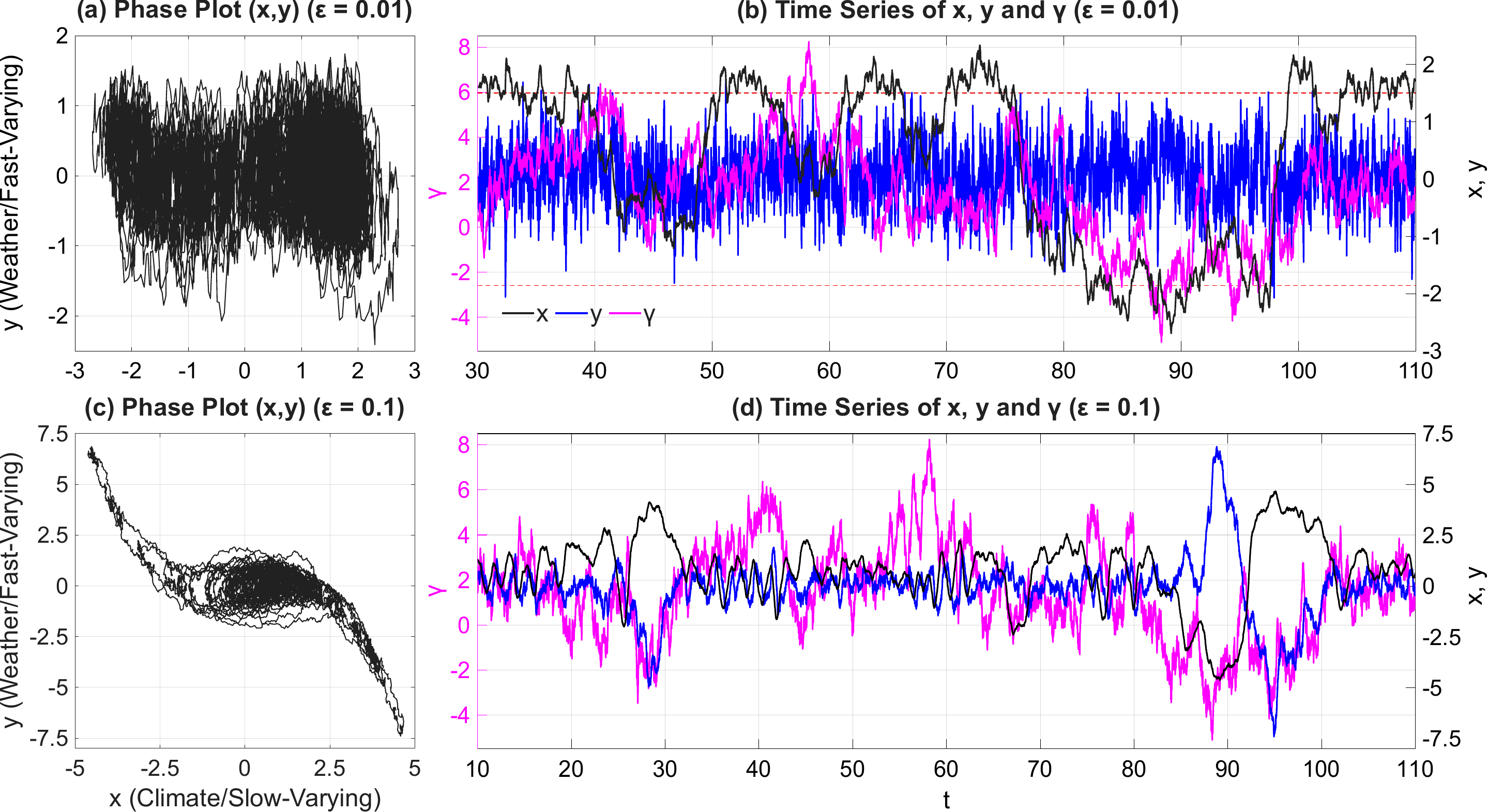}
\caption{Panel (a): Phase plot of $(x(t),y(t))$ for $\varepsilon=0.01$. Panel (b): Time series for $\varepsilon=0.01$. The red dashed horizontal lines indicate the approximate mean stable states for $x$ in each regime ($x\approx 1.5$ and $x\approx -2$). Panel (c--d): Same for $\varepsilon=0.1$. Time windows: $t\in[30,110]$ ($\varepsilon=0.01$), $t\in[10,110]$ ($\varepsilon=0.1$). The random number seeds are fixed, therefore the time series of $\gamma$ in these two simulations is the same.}
\label{fig:climate_tip_fig_1}
\end{figure*}

Figure~\ref{fig:climate_tip_fig_1} reveals distinct behaviors in the two regimes with different $\varepsilon$ values. When $\varepsilon=0.01$, a bistable slow mode $x$ is observed with approximate mean stable states at $x\approx 1.5$ and $x\approx -2$, and state transitions at $t\in[76,82]$ ($x\approx 1.5 \to -2$) and $t\in[97.5,100]$ (reverse). A near-tip instability also occurs during $t\in[40,50]$. Transitions and near-tip instabilities coincide with (near) sign changes of $\gamma$, with $\gamma<0$ especially driving anti-correlated $y$ amplifications that trigger bifurcations in $x$ via the $-\alpha y$ forcing. As such, tips are partly initiated due to the internal variability in \eqref{eq:climate_model2}. On the other hand, when $\varepsilon=0.1$, intermittent oscillations and unstable excursions in $x$ occur for $\gamma<0$ due to the closer $x$-$y$ timescales, which is also evident in near-synchronous oscillations during stable or quiescent periods when $\gamma>0$.

\begin{figure*}[!ht]%
\centering
\makebox[\textwidth][c]{\includegraphics[width=\textwidth]{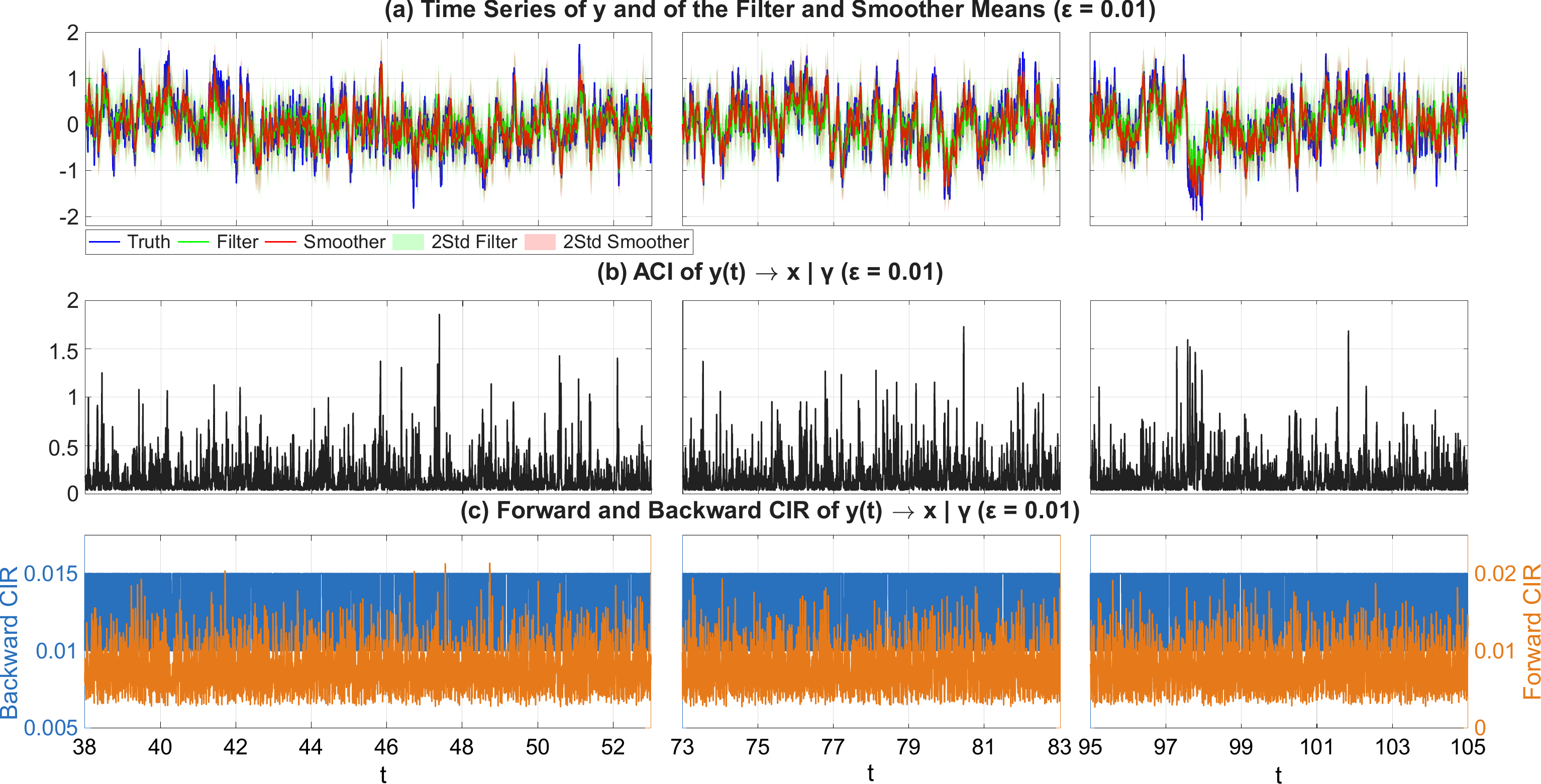}}
\caption{ACI and CIR analysis of $y\rightarrow x|\gamma$ for $\varepsilon=0.01$. Panel (a): Time series with filter/smoother distributions. Panel (b): ACI metric. Panel (c): Forward (orange) and backward (blue) CIR. Columns show $t\in[38,53]$, $t\in[73,83]$, and $t\in[95,105]$, respectively.}
\label{fig:climate_tip_fig_2}
\end{figure*}

\begin{figure*}[!ht]%
\centering
\makebox[\textwidth][c]{\includegraphics[width=\textwidth]{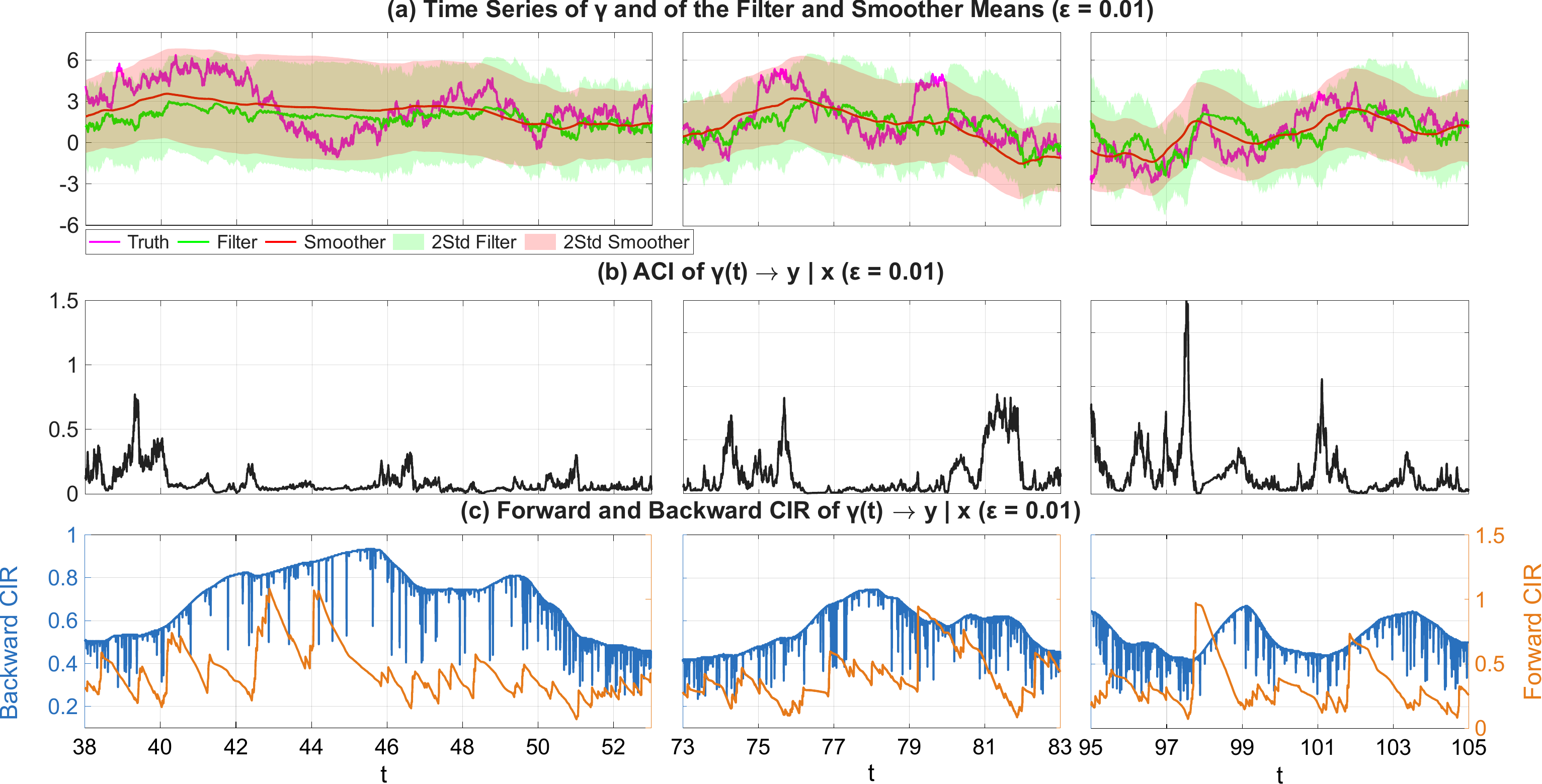}}
\caption{Same as Figure~\ref{fig:climate_tip_fig_2}, but for $\gamma\rightarrow y|x$.}
\label{fig:climate_tip_fig_3}
\end{figure*}

Figures~\ref{fig:climate_tip_fig_2} and~\ref{fig:climate_tip_fig_3} present the ACI and CIR analyses for $\varepsilon=0.01$, numerically validating both the physical intuition behind \eqref{eq:climate_model} and Theorem \ref{thm:obj_backward_CIR_linear} while clarifying causal mechanisms during regime transitions. For $y\rightarrow x|\gamma$ (Figure \ref{fig:climate_tip_fig_2}(b)), significant ACI peaks correlate with tip generation and duration, particularly during the build-up phases of the first and second (return) transitions at $t\in[79,80]$ and $t\approx 98$, respectively. Additional significant local ACI maxima occur during the near-tip instability around $t\in[40,50]$. These temporal patterns reflect $y$'s role as a constant-feedback noisy forcing, where peaks coincide with strong $y$ values, while rapid fluctuations stem from slow-fast scale separation and its internal stochastic variability.

The ACI metric for $\gamma\rightarrow y|x$ (Figure \ref{fig:climate_tip_fig_3}(b)) shows significant local maxima preceding both tips ($t\in[74,76]$ and $t\in[96,98]$) and the near-tip instability ($t\approx 39$) well in advance. Notably, it also peaks immediately after regime establishment ($t\approx 82$ and $t\approx 102$) but does not do so for after the near-tip instability; this indicates how, for this instance, it is the strong negative $y$ values that help terminate the near-tip instability. However, during intermediate transition phases, this ACI metric drops to near zero, suggesting that it is $y$ driving tip maintenance and potential nullification, while $\gamma$'s influence recedes temporarily.

Both forward and backward CIRs for $y\rightarrow x|\gamma$ (Figure \ref{fig:climate_tip_fig_2}(c)) remain short due to the fast timescale of $y$, indicating instantaneous ``short-memory'' causal impact. The backward CIR's uniformity in observational time, oscillating between $0.01$ and $0.015$, confirms Theorem~\ref{thm:obj_backward_CIR_linear}, with equilibrium statistics reached by $T=38$ (and much earlier given the CGNS structure). In contrast, while $\gamma\rightarrow y|x$ exhibits smaller ACI metric values it consistently has much larger CIRs. Sudden forward CIR peaks align with gradual ACI metric increases, with the slow future temporal influence decay  reflecting extended causal horizons, while backward CIR effectively attributes tips to past internal variability, though occasional operationally inflated values (see Remarks \ref{rmk:forward_CIR_properties} and \ref{rmk:backward_CIR_properties}) occur during transitional periods dominated by weather fluctuations and a gradual increase of ACI toward a near-zero local maximum.

These analyses reveal distinct causal patterns for each significant event. The near-tip instability ($t\in[40,50]$) is jointly initiated by bifurcation and noise mechanisms, with comparable pre-excursion ACI values for both relationships. During the instability itself, the ACI metric of $\gamma\rightarrow y|x$ drops to near zero while that of $y\rightarrow x|\gamma$ maintains multiple local maxima, indicating resolution primarily through $y$'s bifurcation-parameter behavior. The first tip ($t\in[76,82]$) is predominantly noise-induced with bifurcation contributions, as both ACI metrics peak beforehand, though $y$ dominates the transition dynamics until $\gamma$ changes sign and becomes strongly negative so as to maintain the negative stable state of $x\approx -2$; post-stabilization, $\gamma\rightarrow y|x$ regains significance with strong ACI local maxima and long CIRs around $t\in[80,82]$, consistent with $\gamma<0$. The second (return) tip ($t\in[97.5,100]$), happening at a much faster timescale, is primarily bifurcation-initiated, reflecting the most significant ACI metric local maxima for $y\rightarrow x|\gamma$ then, followed closely after by significant ACI and CIR values for $\gamma\rightarrow y|x$ when $\gamma$ starts to increase and become positive, showcasing how the noise-induced $\gamma$-feedback aids in the transition to the positive stable state of $x\approx 1.5$ and subsequent maintenance through internal variability processes.

\begin{figure*}[!ht]%
\centering
\makebox[\textwidth][c]{\includegraphics[width=\textwidth]{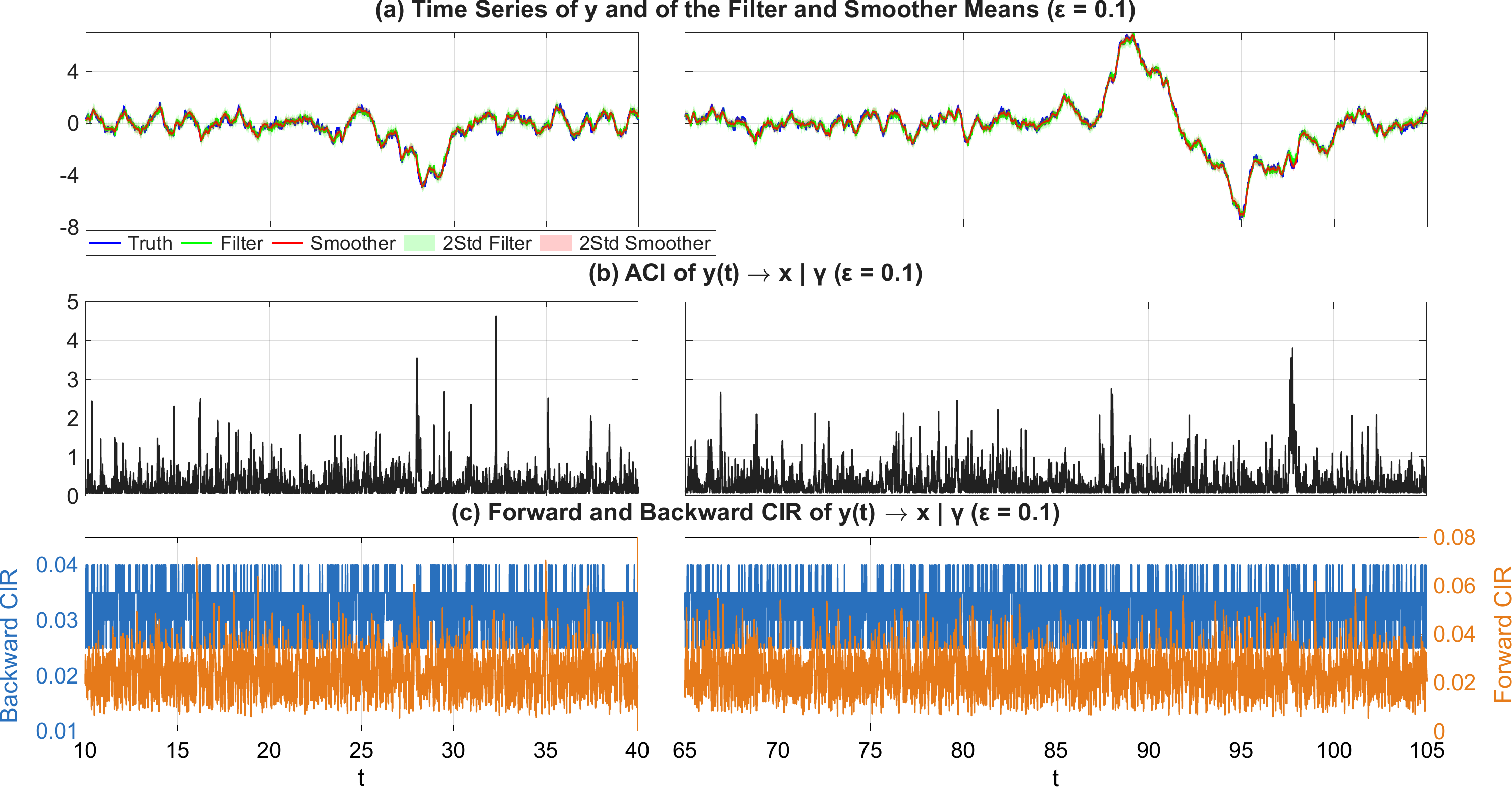}}
\caption{Same as Figure~\ref{fig:climate_tip_fig_2}, but for $\varepsilon=0.1$. Time windows: $t\in[10,40]$ and $[65,105]$.}
\label{fig:climate_tip_fig_4}
\end{figure*}

\begin{figure*}[!ht]%
\centering
\makebox[\textwidth][c]{\includegraphics[width=\textwidth]{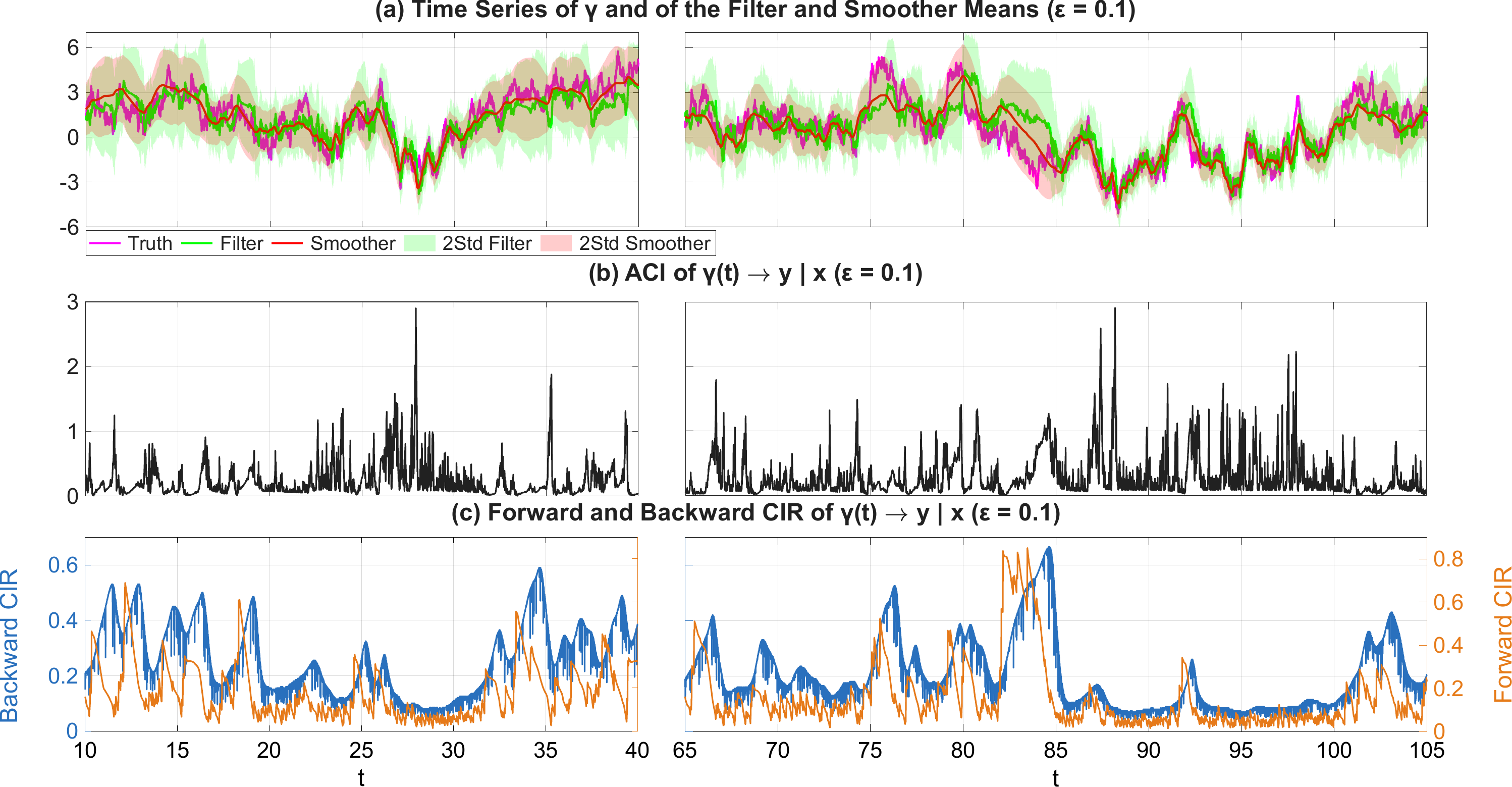}}
\caption{Same as Figure~\ref{fig:climate_tip_fig_4}, but for $\gamma\rightarrow y|x$.}
\label{fig:climate_tip_fig_5}
\end{figure*}

Figures~\ref{fig:climate_tip_fig_4} and~\ref{fig:climate_tip_fig_5} present the ACI and CIR analyses for $\varepsilon=0.1$. In this regime, $x$ exhibits regular oscillations when $\gamma>0$ but becomes more irregular and unstable when $\gamma<0$. Correspondingly, $y$ behaves similarly but with a slight delay due to the reduced timescale separation. Notably, $x$ and $y$ become positively correlated for $\gamma>0$ but anti-correlated for $\gamma<0$, as dictated by the dynamics in \eqref{eq:climate_model1}--\eqref{eq:climate_model2}.

Both ACI metrics show enhanced amplitudes for $\varepsilon=0.1$ throughout time, with $y\rightarrow x|\gamma$ also exhibiting particularly larger CIR values that reflect the reduced timescale separation between $x$ and $y$. When $\gamma$ changes sign and enters strongly negative regimes, it becomes the dominant causal driver (Figure \ref{fig:climate_tip_fig_5}(a--b)), inducing unstable oscillatory excursions in both variables through internal variability. However, the corresponding short CIRs of $\gamma\rightarrow y|x$ during these unstable periods indicate that $\gamma$'s causal influence remains temporally confined after it has already changed sign, affecting instantaneous behavior without extended temporal impact. Notably, prior to the return to stable oscillations when $\gamma$ becomes positive, particularly around $t=32$ and $t=97$, the ACI for $y\rightarrow x|\gamma$ displays concurrent strong local ACI peaks with significant dispersion components (see signal--dispersion decomposition in Appendix~\ref{sec:Tool_Forward_CIR} and Appendix~\ref{sec:Tool_Backward_CIR}). As such, these peaks serve as operational precursors, confirming that internal variability, rather than mean bifurcation dynamics, governs the onset and demise of unstable phases. Once $x$ stabilizes around $x\approx 2$ with regular oscillations when $\gamma>0$, the system becomes predominantly bifurcation-driven, exhibiting correlated but delayed responses through the $-\alpha y$ and $\gamma x$ feedback loop.

The ACI framework proves particularly effective for identifying noise-induced instabilities with $\varepsilon=0.1$, even in the absence of clear early warning signals. The instability and excursion in $x$ with anti-correlating $y$ emerging around $t=85$ demonstrates this capability, with the forward CIR peak at $t=82.5$ providing advance prediction of this significant event, while the strong backward CIRs slightly after, during the initial stages of the unstable phase at around $t\in[83,85]$, correctly attribute the event to $\gamma$'s sign change and evolution toward significant negative values. This behavior, while also present for $\varepsilon=0.01$, becomes more pronounced with the larger $\varepsilon$ as it better resolves this underlying causal structure. The CIR for $y\rightarrow x|\gamma$ also exhibits more intricate temporal structure compared to the $\varepsilon=0.01$ case, consistent with the closer $x$-$y$ timescales, while Theorem~\ref{thm:obj_backward_CIR_linear} remains valid with intermittent backward CIR oscillations between the constants $0.025$ and $0.04$ through intermediate transitions.

Although not shown here for brevity's sake, during quiescent phases of the system, defined by the stable evolution of $x$ (e.g., around $[55,70]$ and $[85,95]$), we can observe that the ACI metric and CIRs of $y\rightarrow x|\gamma$ showcase a uniform baseline behavior, with slight relative increases in ACI values occurring at intermittent slight instabilities of $x$ away from its mean stable states when those have been caused, at least partly, due to instantaneous $y$-bifurcation feedback effects (since intermittent fluctuations that lead to near sign changes in $\gamma$ can also partly trigger these instabilities through noise-induced internal variability). These observations extend to $\gamma \rightarrow y|x$ as well, where this baseline behavior is more pronounced and reflected through near-zero or negligible ACI values and uniform CIRs (although caution is warranted in interpreting these CIRs when the caveats described in Remarks \ref{rmk:caveat_forw_CIR} and \ref{rmk:caveat_back_CIR} occur in the respective ACI metric). Likewise, intermittent peaks in the ACI metric and CIRs can be observed when $\gamma$ suddenly moves toward zero and nearly changes sign, where such fluctuations lead to internal variability that can trigger near-tip instabilities in $x$, as aforementioned.

\subsection{Impact of Resolving Non-Target Variables in the CIR Metrics of a Multiscale Model for Atmospheric Variability} \label{sec:Multiscale}

We consider a system that includes quadratic terms which preserve the system’s energy through the nonlinear operator, multiscale spatiotemporal features, and correlated additive and multiplicative noise. These features enable an efficient search of the non-Gaussian parameter space with far fewer degrees of freedom \cite{majda2009normal, majda2008applied}:
\begin{subequations} \label{eq:multiscale}
    \begin{gather}
        \begin{split}
            \frac{\mathrm{d}x_1}{\rmd t}=a_1x_1-c_1x_1^3-x_2(M+M_{1}x_1+M_{2}x_2)+I_{11}x_1y_1+I_{12}x_1y_2+L_{11}y_1+L_{12}y_2\\
            +f^x_1(t)+\sigma_{x_1}\dot{W}_{x_1}+\frac{\sigma_{y_1}}{\gamma_1}(L_{11}-I_{11}x_1)\dot{W}_{y_{1}}+\frac{\sigma_{y_2}}{\gamma_2}(L_{12}-I_{12}x_1)\dot{W}_{y_{2}},
        \end{split}  \label{eq:multiscale1} \\
        \begin{split}
            \frac{\mathrm{d}x_2}{\rmd t}=-c_2x_2+x_1(M+M_{1}x_1+M_{2}x_2)+I_{21}x_2y_1+I_{22}x_2y_2+L_{21}y_1+L_{22}y_2\\
            +f^x_2(t)+\sigma_{x_2}\mathrm{d}\dot{W}_{x_2}+\frac{\sigma_{y_1}}{\gamma_1}(L_{21}-I_{21}x_2)\dot{W}_{y_{1}}+\frac{\sigma_{y_2}}{\gamma_2}(L_{22}-I_{22}x_2)\dot{W}_{y_{2}},
        \end{split}  \label{eq:multiscale2} \\
        \frac{\mathrm{d}y_1}{\rmd t}=-\frac{\gamma_1}{\varepsilon}y_1-L_{11}x_1-L_{21}x_2+Ny_2-I_{11}x_1^2-I_{21}x_2^2+f^y_1(t)+\frac{\sigma_{y_1}}{\sqrt{\varepsilon}}\dot{W}_{y_{1}}, \label{eq:multiscale3}\\
        \frac{\mathrm{d}y_2}{\rmd t}=-\frac{\gamma_2}{\varepsilon}y_2-L_{12}x_1-L_{22}x_2-Ny_1-I_{12}x_1^2-I_{22}x_2^2+f^y_2(t)+\frac{\sigma_{y_2}}{\sqrt{\varepsilon}}\dot{W}_{y_{2}}. \label{eq:multiscale4}
    \end{gather}
\end{subequations}
The model parameters used in this study are: $a_1 = 1$, $c_1 = 1/3$, $M = 0.5$, $M_1 = 0.5$, $I_{11} = 0.6$, $I_{12} = 0$, $L_{11} = 1$, $L_{12} = 0$, $f_1^x(t) = 0$, ${\sigma_{x_1} = 0.15}$, ${c_2 = 0.4}$, $M_2 = -1.5$, $I_{21} = 0$, $I_{22} = 2$, $L_{21} = 0$, $L_{22} = 1.5$, \linebreak ${f_2^x(t) = 4 + \frac{1}{2}\sin(2\pi t/36)}$, $\sigma_{x_2} = 0.3$, $\gamma_1 = 0.5$, $\varepsilon = 0.1$, $N = 4$, $f_1^y(t) = 1$, $\sigma_{y_1} = 1$, $\gamma_2 = 1.2$, $f_2^y(t) = -1$, and $\sigma_{y_2} = 2$.
Under these model parameter values, the causal network induced by this four-dimensional system is the following:

\begin{figure}[!ht]
\centering
\resizebox{0.3\textwidth}{!}{%
\begin{circuitikz}
\tikzstyle{every node}=[font=\Huge]
\draw [ line width=2pt ] (4,14.75) circle (0.75cm);
\node [font=\Huge] at (4,9.25) {$y_1$};
\draw [ line width=2pt ] (4,9.25) circle (0.75cm);
\node [font=\Huge] at (4,14.75) {$x_1$};
\draw [ line width=2pt ] (9.5,14.75) circle (0.75cm);
\node [font=\Huge] at (9.5,14.75) {$x_2$};

\draw [ line width=2pt ] (9.5,9.25) circle (0.75cm);
\draw [line width=2pt, <->, >=Stealth] (4,14) -- (4,10);
\node [font=\LARGE] at (6.75,11.75) {};
\draw [line width=2pt, <->, >=Stealth] (9.5,14) -- (9.5,10);
\draw [line width=2pt, <->, >=Stealth] (8.75,14.75) -- (4.75,14.75);
\draw [line width=2pt, <->, >=Stealth] (8.75,9.25) -- (4.75,9.25);
\draw [line width=1.5pt, <->, >=Stealth] (9,9.75) -- (4.5,14.25);
\draw [line width=1.5pt, <->, >=Stealth] (4.5,9.75) -- (9,14.25);
\node [font=\Huge] at (9.5,9.25) {$y_2$};
\draw [line width=1.5pt, short] (7.75,13.5) -- (8.5,13.25);
\draw [line width=1.5pt, short] (8.5,13.5) -- (7.75,13.25);
\draw [line width=1.5pt, short] (5.75,13.5) -- (5,13.25);
\draw [line width=1.5pt, short] (5,13.5) -- (5.75,13.25);
\end{circuitikz}
}%
\label{fig:network}
\end{figure}

The system in \eqref{eq:multiscale} is a reduced-order model for large-scale geophysical flow \cite{majda2006nonlinear, ghil2012topics}, often derived via a stochastic-mode reduction on general circulation models \cite{majda1999models, majda2001mathematical}. It distinguishes between the resolved slow atmospheric or environmental variables $\vx=(x_1,x_2)^\mathtt{T}$ and fast weather components $\vy=(y_1,y_2)^\mathtt{T}$ through the timescale separation governed by $0<\varepsilon<1$, which introduces variations in system memory and autocorrelation.

The model captures key physical processes. Quadratic nonlinearities enforce energy conservation while inducing intermittent instabilities in the atmospheric components via antidamping. The linear operator generates multiscale structures typical of turbulent flows, decomposable into a skew-symmetric part mimicking Coriolis effects and topographic Rossby wave propagation, and a negative-definite component providing stochastic stability through dissipative processes like surface drag. In the weather dynamics, oscillatory structures arise from terms like $+Ny_2$ and $-Ny_1$, which dominate for large $N$ while atmospheric variables remain slow if weather feedbacks are rapidly damped. Forcings are scale-dependent: Large-scale terms $f_1^x,f_2^x$ represent slow external inputs such as decadal oscillations \cite{vallis2017atmospheric, mantua2002pacific}, while small-scale terms $f_1^y,f_2^y$ capture faster intraseasonal variability like Madden--Julian oscillation effects \cite{thual2014stochastic}.

The goal of this study is to investigate how resolving ancillary variables clarifies the underlying causal structure. By conditioning on $(x_2,y_2)$ when assessing $y_1\rightarrow x_1$ and on $(x_1,y_1)$ for $y_2\rightarrow x_2$ through conditional ACI, we decompose the joint relationship $(y_1,y_2)\rightarrow (x_1,x_2)$ to identify dominant causal drivers and their temporal influence.

\begin{figure*}[!ht]%
\centering
\includegraphics[width=0.79\textwidth]{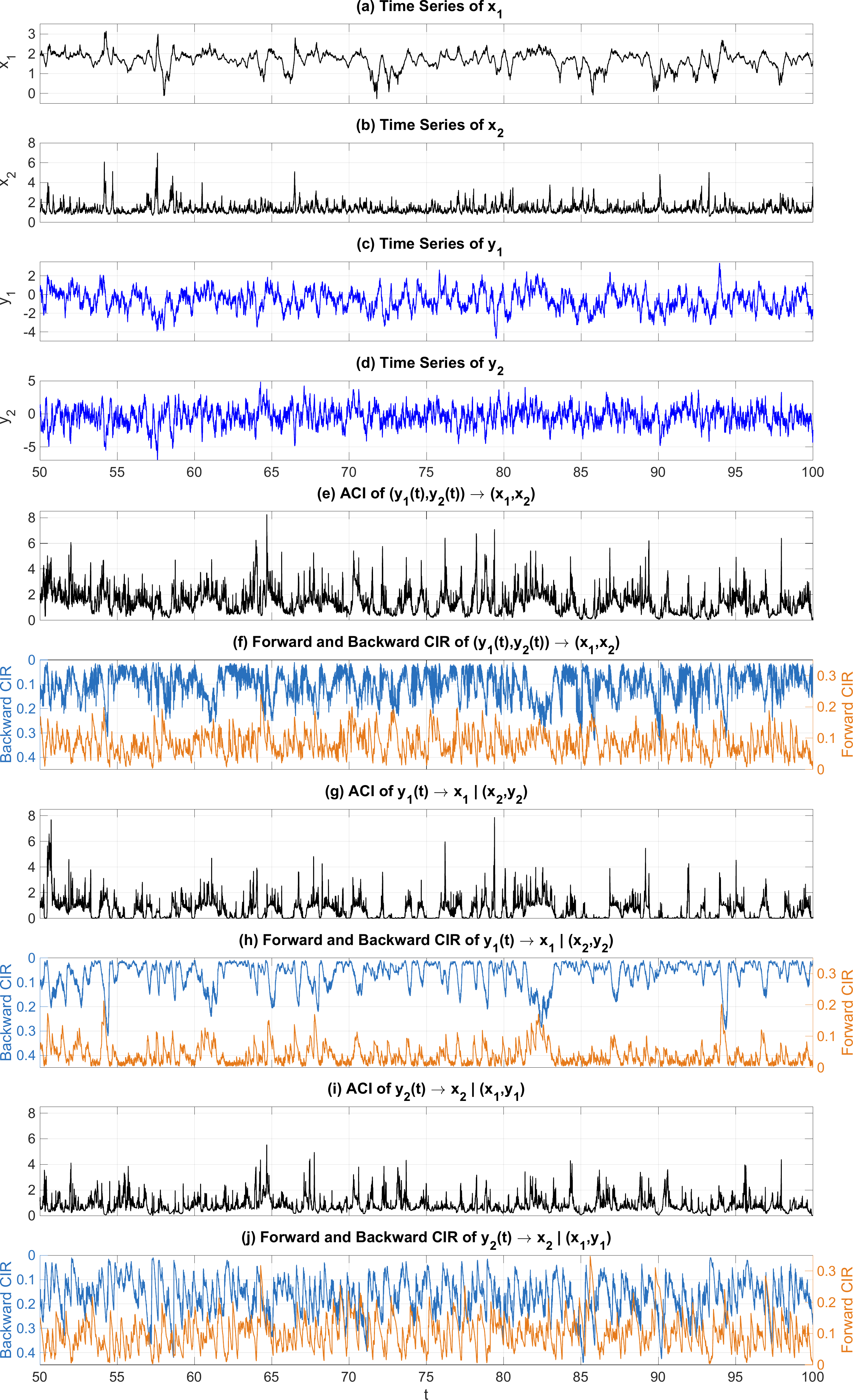}
\caption{Panels (a--d): Time series of $x_1$, $x_2$, $y_1$, and $y_2$. Panel (e): ACI metric used to assess the causal relationship $(y_1,y_2)\rightarrow (x_1,x_2)$ (see \eqref{eq:RE_filter_smoother}--\eqref{eq:ACI_cause_notation}). Panel (f): (Objective) Forward (orange) and backward (blue) CIR of $(y_1,y_2)\rightarrow (x_1,x_2)$; backward CIRs are plotted on a reverse y-axis for ease of comparison with the forward CIRs and to illustrate their complementary and mathematically dual nature as described in Appendix \ref{sec:CIR_Connections}. Panels (g--h): Same as Panels (e--f), but for the conditional causal relationship $y_1\rightarrow x_1 \big| (x_2,y_2)$ (see \eqref{eq:ACI_cause_cond_notation}--\eqref{eq:filter_smoother_ancillary_inf_uncert}). Panels (i--j): Same as Panels (e--f), but for the conditional causal relationship $y_2\rightarrow x_2 \big| (x_1,y_1)$ (see \eqref{eq:ACI_cause_cond_notation}--\eqref{eq:filter_smoother_ancillary_inf_uncert}). Time window: $t\in[50,100]$.}
\label{fig:multiscale_fig}
\vspace*{-69.477pt}
\end{figure*}

The time series in Figure~\ref{fig:multiscale_fig}(a--d) reveal distinct dynamical behaviors. The variable $x_1$ exhibits intermittent excursions away from its mean stable state $x_1\approx 1.8$ toward a stronger or weaker but unstable state, with a proclivity for the latter, thus showing negative skewness and a slight multimodality due to cubic dissipation, multiplicative and correlating noise, and weak weather-mode feedbacks \cite{majda2009normal, chen2025stochastic}. In contrast, $x_2$ remains strictly positive with significant positive skewness and intermittent instabilities (significant positive bursts away from its mean stable state $x_2\approx 1.4$) due to energy conservation restrictions. The atmosphere-to-weather influence components are asymmetric: $x_2$ exerts stronger control on $y_2$ than $x_1$ does on $y_1$, consistent with $I_{11}<I_{22}$ and $L_{11}<L_{22}$. Consequently, $y_1$ follows a sub-Gaussian distribution while both $y_1$ and $y_2$ showcase a slight negative skewness, with both weather variables evolving faster than their atmospheric counterparts ($\varepsilon=0.1$). All marginals and joint distributions display strong non-Gaussianity due to the intermittent instabilities, regime switches, and multiscale, nonlinear dynamics.

Panels (e--j) demonstrate how resolving ancillary variables affects ACI metrics and CIRs. Conditioning increases CIR lengths relative to the unconditional case (Panels (e--f)), especially for $y_2 \rightarrow x_2|(x_1,y_1)$, which indicates the longer memory effects involved in the physical-constraint of energy conservation where $x_2$ occasionally (depending on the $\vy$ values) needs to amplify to compensate $x_1$'s intermittent rapid and unstable excursions generated either internally, by the weather component $y_1$, or through the multiplicative and correlated noise. Furthermore, conditioning reveals the true causal structure by decomposing when large joint ACI values stem from one marginal relationship while the other remains negligible. Specifically, accounting for $(x_2,y_2)$ in $y_1\rightarrow x_1$ (Panels (g--h)) shows that the weather-to-atmosphere feedback from $y_1$ to $x_1$ predominantly drives the dispersion component of the joint ACI metric (see signal--dispersion decomposition in Appendix~\ref{sec:Tool_Forward_CIR} and Appendix~\ref{sec:Tool_Backward_CIR}), particularly during the intervals around $t\in[50,55]$, $t\in[60,62.5]$, $t\in[80,84]$, and $t\in[93,95]$, where $y_1$'s positive and rapid fluctuations generate $x_1$'s multimodal tendencies through unstable excursions. As expected, these significant ACI peaks with predominant dispersion parts coincide with longer CIRs for $y_1\rightarrow x_1|(x_2,y_2)$ (see Appendix \ref{sec:Tool_Forward_CIR}--\ref{sec:Tool_Backward_CIR} for details). Conversely, conditioning on $(x_1,y_1)$ in $y_2\rightarrow x_2$ (Panels (i--j)) reveals the aforementioned longer CIRs attributed to $y_2$'s weather effects, with the significant CIRs occurring at around $t=57.5$, $t\in[64,65]$, $t\in[85,86]$, and $t=90$. These timings, considering that the observed longer CIRs enable prediction of $x_2$ excursions up to $0.3$ units ahead and their attribution up $0.4$ units into the past to strong negative $y_2$ amplitudes, coincide with when the ACI metric of $y_1\rightarrow x_1|(x_2,y_2)$ drops close to zero while that for $y_2\rightarrow x_2|(x_1,y_1)$ evolves slowly toward a significant local maximum. In terms of the physical state space, these periods correspond to $x_2$'s intermittent instabilities and positive bursts which force $x_1$ into unstable excursions, or the other way around depending on the weather causality driven by $\vy$, so as to preserve the system's (quadratic) energy conservation. We also note that the conditional ACI metric for $y_2\rightarrow x_2|(x_1,y_1)$ never collapses down to zero, indicating persistent uncertainty reduction from smoother solutions that is consistent with the longer CIRs and a nontrivial dispersion component over time. In contrast, the ACI metric of $y_1\rightarrow x_1$ is negligible during $x_1$ excursions driven by the weather $y_1$ effects, as these deterministic events are well-captured by the filter alone due to strong system observability and signal-to-noise ratios at those times.

\clearpage

\thispagestyle{fancy}
\fancyhf{}
\renewcommand{\headrulewidth}{0pt}
\setlength{\headheight}{14.49998pt}
\fancyhead[C]{\thepage}

\clearpage

\subsection{Atmospheric Blocking Mechanisms in an Equatorial Circulation Model with Seasonality and Flow-Wave Interactions} \label{sec:Lorenz84}

The noisy Lorenz-84 model is a conceptual system describing coarse-grained equatorial atmospheric circulation \cite{vallis2017atmospheric, salmon1998lectures}. Its deterministic component is derived from a Galerkin truncation of the quasi-geostrophic potential vorticity equations \cite{chen2018conditional, van2003baroclinic}. The model captures key features of the Hadley cell and can be extended to include atmosphere-ocean coupling for studying oceanic variability and seasonality \cite{chen2018conditional, roebber1995climate, ferrari2003seasonal}.
The model reads \cite{lorenz1984irregularity}:
\begin{subequations} \label{eq:lorenz84model}
    \begin{align}
        \frac{\mathrm{d}x}{\rmd t}&=-y^2-z^2-a\big(x-f(t)\big)+\sigma_x\dot{W}_x, \label{eq:lorenz84model1}\\
        \frac{\mathrm{d}y}{\rmd t}&=-bxz+xy-y+g+\sigma_y\dot{W}_y, \label{eq:lorenz84model2}\\
        \frac{\mathrm{d}z}{\rmd t}&=bxy+xz-z+\sigma_z\dot{W}_z. \label{eq:lorenz84model3}
    \end{align}
\end{subequations}
The model parameters used for this case study are: $a=1/4$, $f(t)=8+3\cos(2\pi t/73)$, $\sigma_x=0.2$, $b=4$, $g=1$, and $\sigma_y=\sigma_z=0.2$, which follow those originally proposed by Lorenz and define a characteristic time scale of about five days \cite{lorenz1984irregularity, lorenz1990can}. The Lorenz-84 model \eqref{eq:lorenz84model} describes atmospheric wave-mean flow interactions where the unobserved variable $x$ represents the amplitude of the westerly zonal flow, while the observed variables $y$ and $z$ are the cosine and sine phases of a large-scale baroclinic wave. Under a thermal wind balance formulation, $x$ can describe the zonally-averaged meridional temperature gradient \cite{vallis2017atmospheric, majda2006nonlinear}. The quadratic nonlinearities conserve energy and capture wave amplification at the expense of the zonal flow, with wave displacement governed by the advection strength $b$. The Prandtl number $a<1$ allows the zonal flow to dampen slower than waves. External forcings include the seasonal thermal contrast $f(t)=8+3\cos(2\pi t/73)$, which drives the zonal flow and incorporates a one-year period, and the constant asymmetric forcing $g=1$ that represents topographic influences. The model exhibits bistability between two stable equilibria, mimicking transitions between zonal (strong westerly jets) and blocked (amplified waves) atmospheric regimes. This provides a framework for using ACI to study how the jet $x$ influences waves $y$ and $z$ during blocking events.

Below, we employ forward and backward CIRs to quantify the temporal causal extent of the zonal wind current ($x$) on the large-scale vortices ($y$,$z$), mirroring teleconnections where regional Earth process anomalies are related via atmospheric pathways. Forward CIR analysis can be used to identify whether an increasing zonally-averaged meridional temperature gradient can function as a causal precursor to rapidly oscillating waves in the future, while the backward CIR can assess if an observed amplified wave component can be traced back and causally attributed to a past weak jet stream or atmospheric blocking episode driven by poleward heat transport.

\begin{figure*}[!ht]%
\centering
\includegraphics[width=0.85\textwidth]{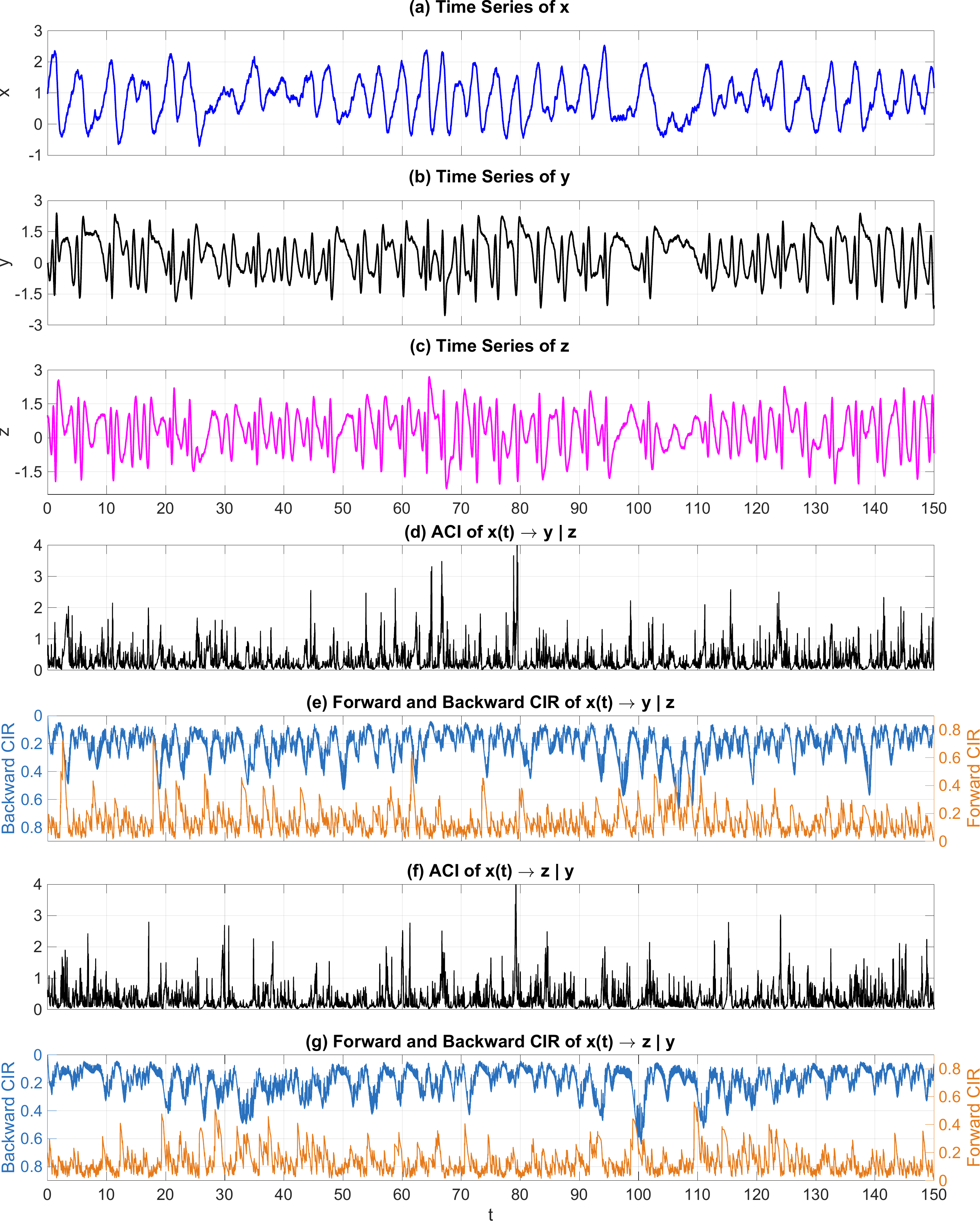}
\caption{Same as Figure~\ref{fig:multiscale_fig}, but for the noisy Lorenz-84 model in \eqref{eq:lorenz84model} and the conditional assimilative causal relationships $x\rightarrow y|z$ (Panels (d--e)) and $x\rightarrow z|y$ (Panels (f--g)). Time window: $t\in[0,150]$.}
\label{fig:lorenz_84_fig}
\end{figure*}

Figure~\ref{fig:lorenz_84_fig}(a--c) reveals chaotic teleconnection patterns: $x$ evolves more slowly than the wave components $y$ and $z$. When $x$ oscillates away from zero, it induces regular wave oscillations ($t\in[15,17.5]$, $t\in[30,50]$, $t\in[110,125]$), whereas near-zero $x$ values (atmospheric blocking) cause protracted wave oscillations ($t\in[7.5,10]$, $t\in[25,30]$, $t\approx50$, $t\in[95,100]$, $t\in[102.5,110]$). During blocking, phase-shifted $y$ and $z$ reach opposite peaks before decaying irregularly due to energy conservation, consistent with poleward heat transport weakening the jet through the vortical term ${-y^2-z^2}$. Wave growth is in time suppressed as weak $x$ enforces linear stability in \eqref{eq:lorenz84model2}--\eqref{eq:lorenz84model3}, eventually leading to zonal flow resurgence.

Analysis of the conditional relationships $x\rightarrow y|z$ (Panels (d--e)) and $x\rightarrow z|y$ (Panels (f--g)) shows $x$ predominantly drives $y$ and $z$ during stable oscillations away from zero, particularly when approaching and then exceeding their antidamping threshold of $x=1$ ($t=45$, $t\in[60,70]$, $t=80$, $t=115$, $t=125$), while the ACI metrics, as expected, drop to near zero during the aforementioned blocking events due to $y$- and $z$-causality dominating the dynamics. The former regime induces rapid $y$ and $z$ oscillations that show significant ACI but short CIRs, indicating $x$ initiates but doesn't maintain instabilities due to energy conservation. The strongest ACI local maxima occur around $t=67.5$, $t=80$, $t=124$, where transitional phases between damped and rapid oscillations confirm the model's core mechanism: Zonal jet damping by significant vortical amplitudes versus wave instability from increasing meridional temperature gradients.

During $x$'s slow irregular oscillations away from zero ($t\in[15,17.5]$, $t\in[30,50]$, $t\in[110,125]$), both ACI and the CIRs show local peaks, suggesting zonal displacement dominates amplification with persistent causal influence. Similar patterns appear in $x\rightarrow y|z$ and $x\rightarrow z|y$, though phase differences from the asymmetric topographic forcing $g$ in $y$ creates discrepancies in the system's observability. During blocking events (especially $t\approx50$, $t\in[95,100]$, $t\in[102.5,110]$), ACI values drop relatively sharply as waves become primary causal drivers, while an observed unblocking is persistently attributed to weak $x$ in the past via the backward CIR, as it generates linear stability. Notable is the extended blocking event during $t\in[100,110]$, which shows the most distinct and discernible patterns. Specifically, diminished zonal gradients can result from heat transport without full blocking, a nuance often missed in physical studies. Throughout, backward CIR attributes weak jets to prior wave amplification, while forward CIR warns of future rapid oscillations during strong jets.

\section{Discussion and Conclusion} \label{sec:Conclusion}

In this paper, we developed mathematically rigorous formulations for quantifying the temporal extent of causal impact within the ACI framework. We introduced CIRs that measure the future period in which a current cause will remain influential (forward CIR) and the past period of causes responsible for a currently observed effect (backward CIR). Theoretically grounded approximations enable efficient computation of these temporal measures and facilitate asymptotic analysis in tractable complex systems. This work advances causal inference beyond static relationships, bridging prediction and attribution of intermittent extreme events and regime transitions. Applications on reduced-order yet physically meaningful models reveal the effectiveness of the framework in identifying causal drivers of tipping points, particularly for those triggered by internal variability which lack conventional early warning signals. The conditional formulation of ACI and the CIRs further enables the identification of overlapping causal influences in high-dimensional multiscale systems with correlated multiplicative noise by resolving interfering or spurious non-target effects. The CIRs also advance the investigation of atmospheric blocking mechanisms through a conceptual equatorial circulation model with seasonal feedbacks and flow-wave interactions.

Future research directions include the following. First, developing methods to identify significant marginal assimilative causal relationships and their associated CIRs in multivariate settings would enhance causal discovery and allow for a more versatile and comprehensive framework. Second, the partial-observability assumption of the ACI framework not only allows for natural causal inference, under the assumption that we only observe subsequent effects which we then trace backward through the underlying dynamical system to their candidate causes as an inverse problem, but also restricts the causal state space to the latent or unobserved subspace, unlike traditional methods, over which posterior PDFs are reconstructed. Therefore, leveraging computationally efficient methods developed in the mature field of Bayesian data assimilation to overcome the curse of dimensionality, including ensemble Kalman--Bucy filters and smoothers where ensemble sizes are often chosen independently of the state dimension \cite{jiang2026continuous}, we can assess the scalability and robustness of ACI and of its forward and backward CIRs to high-dimensional complex systems, especially against other state-of-the-art causal inference methods for instantaneous, time-dependent causality. Third, deriving analytical expressions for conditional ACI and CIRs in general complex nonlinear systems with cross-interacting noise feedbacks would strengthen the theoretical foundations of the framework. Fourth, investigating ACI and the proposed CIR metrics under noise-contaminated observations would improve practical applicability. Finally, assessing the robustness of the framework against model error remains crucial. Such studies could also open new avenues for constructing reduced-order parsimonious models or integrating ACI and its CIRs to digital twin experiments.

\section*{Funding}
The research of N.C. is funded by the Office of Naval Research N00014-24-1-2244 and the Army Research Office W911NF-23-1-0118. M.A. is partially supported as a research assistant under the second grant.

\section*{Author Contributions: CRediT}
\textbf{Marios Andreou:} Conceptualization, Formal analysis, Methodology, Software, Visualization, Writing -- Original draft. \textbf{Nan Chen:} Funding acquisition, Conceptualization, Methodology, Supervision, Writing -- Review \& editing.

\section*{Data Availability Statement}
No external datasets were used in this research; all data in the simulations are model-generated and synthetic.

\section*{Code Availability Statement}
The MATLAB code used in the analyses and to generate the figures in this work can be found in the following GitHub repository: \url{https://github.com/marandmath/FBCIR_code}. The codebase also includes optional implementations for definition-vs-approximation comparison plots between the exact objective forward and backward CIRs (Eqs. \eqref{eq:objective_forward_CIR_length} and \eqref{eq:objective_backward_CIR_length}, respectively) and their respective computationally efficient approximations (Eqs. \eqref{eq:forward_CIR_approximation} and \eqref{eq:obj_backward_CIR_approx}, respectively) over time, for all the case studies presented in this work. Additional optional analyses for all case studies also include the computation and heatmap plotting of the subjective forward and backward CIRs (Eqs. \eqref{eq:subjective_forward_CIR_length} and \eqref{eq:subjective_backward_CIR_length}, respectively) and normalized variants of the forward CIR metric in \eqref{eq:forward_CIR_metric} and complete backward CIR metric in \eqref{eq:backward_CIR_metric}--\eqref{eq:complete_backward_CIR_metric}, with the computational details being provided in Appendix \ref{sec:Tool_CGNS}. Finally, a repository-specific case study of ACI and its CIRs on a multimodal and layered topographic paradigm model with mean-flow interactions for blocking/unblocking regime switching in barotropic flow is also included in the codebase.

{
\appendix

\renewcommand\thefigure{\thesection.\arabic{figure}}
\counterwithin{figure}{section}
\newtheorem{remarkapp}{Remark}[section]
\renewcommand{\thesection}{\Alph{section}}
\makeatletter
\renewcommand\@seccntformat[1]{\appendixname\ \csname the#1\endcsname.\hspace{0.5em}}
\makeatother

\section*{Appendix}

Throughout the Appendix, the same notations and setup as in the main text are used. Furthermore, we posit the following additional remarks and assumptions which we enforce when required:
\begin{itemize}
    \item The relative entropy in \eqref{eq:relative_entropy}, is part of the family of $f$-divergences \cite{polyanskiy2024information, ali1966general}:
    \begin{equation*}
        \mathcal{D}_f(\mathbb{P},\mathbb{Q}):=\int f(\rmd \mathbb{P}/\rmd\mathbb{Q})\rmd\mathbb{Q}.
    \end{equation*}
    In particular, it is induced by the $f(v) = v \log v$ generator function. It is known that an $f$-divergence induced by a lower semi-continuous (LSC) $f$ is itself LSC in the weak topology of the restricted product space of Borel probability measures with $\mathbb{P} \ll \mathbb{Q}$, where $\mathbb{P}$ and $\mathbb{Q}$ are defined over a Hausdorff topological state space \cite{ambrosio2000functions, liero2018optimal}; this product space is equivalent to the direct product of the weak topologies \cite{brezis2011functional}. In other words, if both $\mathbb{P}$ and $\mathbb{Q}$ are dominated by the appropriate Lebesgue measure and enjoy PDFs $p$ and $q$, respectively, then the $f$-divergence is weakly LSC in both the primary density $p$ as well as the reference PDF $q$.

    \item We assume that all backward interpolation PDFs (i.e., the lagged and complete smoother PDFs; see Section~\ref{sec:CIR_Framework}),
    \begin{equation*}
        p\big(\vy(t) \big| \vx(s \leq T')\big),
    \end{equation*}
    are weakly continuous in $T' \in [t, T]$ for each fixed $t \in [0, T]$, as well in $t\in[0,T']$ for each fixed $T'\in[0,T]$. These conditions are necessary because weak continuity, not merely weak LSC, is required for the subsequent arguments, specifically since the composition of LSC functions is not LSC itself necessarily. Note here that this weak continuity condition is equivalent to weak sequential continuity, since we consider at least a reflexive Polish state space (e.g., separable Hilbert space). These weak-continuity properties can be analytically verified for the backward interpolation distributions arising from the class of dynamical systems considered in \eqref{eq:general_CTNDS} \cite{rozovsky2012stochastic}, including the CGNSs introduced in \eqref{eq:CGNS} of Appendix~\ref{sec:Tool_CGNS} \cite{liptser2001statistics}, under mild regularity assumptions.

    \item A Borel measurable function is always Lebesgue measurable in our adopted setup.
\end{itemize}

\section{Conditional ACI: Causality in the Presence of Non-Target Variables} \label{sec:Conditional_ACI}

Let $\vx=(\vx_{\text{A}},\vx_{\text{B}})$ denote the observed variables and $\vy$ the unobserved ones. The conditional variation of ACI developed in  \cite{andreou2026assimilative}, analogous to \eqref{eq:RE_filter_smoother}--\eqref{eq:ACI_cause_notation}, assesses whether the conditional assimilative causal relationship
\begin{equation} \label{eq:ACI_cause_cond_notation}
	\vyt\rightarrow\vx_{\text{A}}\big| \vx_{\text{B}}
\end{equation}
holds at $t\in[0,T]$. In \eqref{eq:ACI_cause_cond_notation}, the non-target variables $\vx_{\text{B}}$ can play the role of confounders, mediators, moderators, colliders, or instrumental variables. In the conditional ACI framework, \eqref{eq:ACI_cause_cond_notation} is established if \cite{andreou2026assimilative}:
\begin{equation} \label{eq:RE_filter_smoother_general}
	\mathcal{P}\big(p^{\text{\normalfont{s}}|\vx_{\text{B}}}_t(\vy|\vx_{\text{A}}), p^{\text{\normalfont{f}}|\vx_{\text{B}}}_t(\vy|\vx_{\text{A}})\big) > 0,
\end{equation}
where
\begin{subequations} \label{eq:filter_smoother_ancillary_inf_uncert}
	\begin{align}
		p^{\text{\normalfont{s}}|\vx_{\text{B}}}_t(\vy|\vx_{\text{A}}) &:=\lim_{\mathrm{Var}(\vx_{\text{B}})\to+\infty} p_t^{\text{\normalfont{s}}}(\vy|\vx), \label{eq:filter_smoother_ancillary_inf_uncert1} \\
		p^{\text{\normalfont{f}}|\vx_{\text{B}}}_t(\vy|\vx_{\text{A}}) &:= \lim_{\mathrm{Var}(\vx_{\text{B}})\to+\infty} p_t^{\text{\normalfont{f}}}(\vy|\vx), \label{eq:filter_smoother_ancillary_inf_uncert2}
	\end{align}
\end{subequations}
are the smoother- and filter-based PDFs of $\vyt$ obtained by assigning infinite uncertainty to $\vx_{\text{B}}$'s marginal likelihood during the analysis stage of Bayesian data assimilation when calculating the smoother and filter posterior PDFs, respectively. 

Recall that in ACI, causality is interpreted as an inverse problem in Bayesian data assimilation, where causal influence is encoded through the smoother-to-filter information gain from the future observations of the potential subsequent effect when we trace them back to their candidate latent cause to be state-estimated. Therefore, the formal limits in \eqref{eq:filter_smoother_ancillary_inf_uncert} represent a mathematical shorthand for the prohibition of $\vx_{\normalfont{\text{B}}}$ from spuriously affecting the Bayesian update of $\vyt$. This ensures, unlike marginalizing or conditioning on $\vx_{\text{B}}$, that any posterior uncertainty reduction in the smoother from the filter is provided solely from the future observations of $\vx_{\normalfont{\text{A}}}$ so as to assess whether the conditional causal relationship $\vyt\rightarrow \vx_{\normalfont{\text{A}}}|\vx_{\normalfont{\text{B}}}$ holds, while maintaining model integrity by still letting $\vx_{\normalfont{\text{B}}}$ naturally be part of the underlying stochastic dynamics \cite{andreou2026assimilative}. This is possible because the influence or weight from the observations on the posterior state update of $\vyt$ is determined by the product of the appropriate Kalman gain operator and the observational innovation, where the former is proportional to the inverse of the observational uncertainty (see \eqref{eq:general_CTNDS}) \cite{andreou2026assimilative}:
\begin{equation} \label{eq:inv_observational_uncertainty}
    \big(\ms_1^\vx(\ms_1^\vx)^\mathtt{T}+\ms_2^\vx(\ms_2^\vx)^\mathtt{T}\big)^{-1}.
\end{equation}

Practically, there are various operational shortcuts to simulate this limiting behavior without actually going through the trouble of empirically calculating the limits in \eqref{eq:filter_smoother_ancillary_inf_uncert}. Specifically, this conditional ACI argument is essentially equivalent to setting the elements corresponding to the non-target variables $\vx_{\normalfont{\text{B}}}$ in \eqref{eq:inv_observational_uncertainty} down to zero during state estimation, leading to the aforementioned modified posterior PDFs. This is analogous to setting the marginal likelihood uncertainty of $\vx_{\normalfont{\text{B}}}$ to infinity, since the diagonal elements of the observational precision matrix are the reciprocals of the marginal likelihood variances of the observed variables, so $1/\mathrm{Var}(x_{\normalfont{\text{B}},i})\to 0$ for each $i$ coordinate, while the off-diagonal elements correspond to the cross-correlations between the observed variables $\vx_{\normalfont{\text{A}}}$ and $\vx_{\normalfont{\text{B}}}$, which are also set to zero in this limit since $\vx_{\normalfont{\text{B}}}$ can be thought of as an analog of a diffuse or flat Bayesian prior in this limiting regime. Notable is that for the conditional Gaussian nonlinear systems (see Appendix \ref{sec:Tool_CGNS}) studied in the original ACI paper and in this work \cite{andreou2026assimilative}, as well as for the general partially-observed dynamical system in Eqs. (1a)--(1b) when using an appropriate ensemble Kalman--Bucy filter and smoother \cite{jiang2026continuous}, this proposed limiting approach corresponds to conducting state estimation through an equivalent reduced-order dynamical system for $(\vx_{\normalfont{\text{A}}},\vy)$, about the $\vy | \vx_{\normalfont{\text{A}}}$ posterior, where $\vx_{\normalfont{\text{B}}}$ is reduced to a deterministic process defined by its observed values/time series, analogous to a control or input term, which are used to calculate the state-dependent model parameters of the dynamics in the filter and smoother state estimation equations.

\section{Mathematical Details and Schematic Illustration of the Forward CIR Framework} \label{sec:Forward_CIR_Details}

We start this appendix by presenting a series of remarks outlining the mathematical properties and operational details concerning the forward CIR metric and its associated subjective and objective lengths.

\begin{remarkapp}[Measurability of the Forward CIR Metric] \label{rmk:forw_CIR_measurability}
    Consider the forward CIR metric from \eqref{eq:forward_CIR_metric}:
    \begin{equation*}
        \delta^{\text{\normalfont{f}}}(\tau;t):=\delta(t, t+\tau(T-t)),
    \end{equation*}
    for $\tau\in\mathrm{I}=[0,1]$ and arbitrary $t\in[0,T]$. Then, as the composition between a weakly LSC function (relative entropy) and weakly continuous functions (backward interpolation PDFs and $h(\tau)=t+\tau(T-t)$), $\delta^{\text{\normalfont{f}}}(\tau;t)$ is a weakly LSC function on $\mathrm{I}$ for each $t\in[0,T]$. This weak lower semi-continuity implies that $\delta^{\text{\normalfont{f}}}(\tau;t)$ is also weakly measurable \cite{diestel1977vector, hytnen2016analysis}. One way to establish strong measurability now is to prove that it is also almost surely separably valued as to apply Pettis' measurability theorem \cite{pettis1938integration}, however this step is unnecessary in our setting. For separable spaces, weak and strong measurability coincide, and the space of Borel probability measures defined over a Polish state space is itself Polish, as established by Prokhorov's theorem \cite{van2003probability, kechris1995classical, parthasarathy2005probability}. As such, the forward CIR metric defined in \eqref{eq:forward_CIR_metric} is a measurable function of $\tau \in \mathrm{I}$ for each $t \in [0, T]$.
\end{remarkapp}

\begin{remarkapp}[Temporal Profile and Interpretations of the Forward CIR Metric] \label{rmk:forward_CIR_properties}
The key temporal properties of the forward CIR metric are:
\begin{itemize}
    \item At $\tau=0$, $\delta^{\text{\normalfont{f}}}(0;t)=\delta(t, t)=\mathcal{P}\big(p_t^{\text{\normalfont{s}}}(\vy|\vx),p_t^{\text{\normalfont{f}}}(\vy|\vx)\big)$ equals the ACI metric assessing $\vyt\rightarrow\vx$ (see \eqref{eq:RE_filter_smoother}--\eqref{eq:ACI_cause_notation}). In chaotic systems, $M^{\text{\normalfont{f}}}(t)$ typically occurs at $\tau=0$ or shortly after, reflecting the fact that causal influence is strongest immediately following the cause.

    \item For $\tau\in(0,1)$, $\delta^{\text{\normalfont{f}}}(\tau;t)$ measures the information deficit from incorporating only observations up to $T'=h(\tau)=t+\tau(T-t)\leq T$ rather than $T$. Therefore, the subinterval in $\mathrm{I}$ over which $\delta^{\text{\normalfont{f}}}(\tau;t)$ is large indicates significant future causal influence, as substantial information gain comes from including the future observations in $(h(\tau),T]$.

    \item At $\tau=1$, $\delta^{\text{\normalfont{f}}}(1;t)=0$ by the positive-definiteness of the relative entropy \cite{cai2002mathematical}.
\end{itemize}
\end{remarkapp}

\begin{remarkapp}[An Inequality for the Subjective Forward CIR Length] \label{rmk:subj_forw_CIR_lebesgue}
The subjective forward CIR length overestimates the measurable persistence of $\vyt$'s causal influence on $\vx$'s future states in $[t,T]$, since it satisfies
\begin{equation*}
    \overset{\sim}{\uptau}^{\,\text{\normalfont{f}}}(t,\varepsilon)\geq (T-t)\lambda_{\mathrm{I}}\big(\big\{\tau\in\mathrm{I} : \delta^{\text{\normalfont{f}}}(\tau;t) > \varepsilon\big\}\big),
\end{equation*}
where $\lambda_{\mathrm{I}}$ is the Lebesgue measure on $\mathrm{I}$.
\end{remarkapp}

\begin{remarkapp}[Operational Details of the Objective Forward CIR] \label{rmk:caveat_forw_CIR}
When $M^{\text{\normalfont{f}}}(t)\approx 0$ (weak evidence for $\vyt\to\vx$ assimilative causality), $\uptau^{\,\text{\normalfont{f}}}(t)$ and its computationally efficient (lower-bound) approximation, $\uptau^{\,\text{\normalfont{f}}}_{\text{approx}}(t)$, may yield an inflated CIR length if $\delta^{\text{\normalfont{f}}}(\tau;t)$ slowly increases toward its near-zero maximum $M^{\text{\normalfont{f}}}(t)$ as $\tau$ decreases toward the neighborhood of $\tau=0$. Such values typically coincide with small or near-zero local maxima of the ACI metric in \eqref{eq:RE_filter_smoother} and should therefore be interpreted jointly with it so as to avoid misrepresentation of temporal causal impact. Numerically, to avoid such operationally-inflated values due to floating-point imprecision, if $M^{\text{\normalfont{f}}}(t)<\zeta$ for some small threshold $\zeta>0$ (e.g., $\zeta=10^{-5}$), we can set $\uptau^{\,\text{\normalfont{f}}}(t)=\uptau^{\,\text{\normalfont{f}}}_{\text{approx}}(t)=0$, which will necessarily coincide with $\overset{\sim}{\uptau}^{\,\text{\normalfont{f}}}(t,\varepsilon)\approx0$ due to the adopted convention of $\overset{\sim}{\uptau}^{\,\text{\normalfont{f}}}(t,\varepsilon)=0$ when $\varepsilon\geq M^{\text{\normalfont{f}}}(t)\approx 0$.
\end{remarkapp}

Figure~\ref{fig:details_FCIR} illustrates the underlying mechanisms of the ACI-based forward CIR theory developed in this work through an illustrative extreme-event scenario (specifically, this analysis corresponds to the example simulation depicted in Panel (II) of Figure~\ref{fig:schematic_diagram}). We focus on a time $t$ during the onset of an extreme event in $\vx$ that is being triggered by $\vyt$. Both the objective and subjective forward CIR formulations are shown. To interpret these results, note that the subjective forward CIR length in \eqref{eq:subjective_forward_CIR_length} is an affine transformation of the generalized inverse of the forward CIR metric in \eqref{eq:forward_CIR_metric} \cite{feng2012note, embrechts2013note, kampke2014generalized}. For the extreme event being generated by $\vyt$, the forward CIR metric fails to be a nonincreasing function for some lagged observational times $T'\in[t,T]$. Consequently, its generalized inverse and by extension the subjective forward CIR length are multivalued functions at the $\varepsilon$ values the monotonicity of the forward CIR metric is first violated in $T'$. This demonstrates the strong need for objective CIR measurements, and in turn also validates that the computationally efficient approximation in \eqref{eq:forward_CIR_approximation} is a strict underestimate of the objective forward CIR in such cases: $\uptau^{\,\text{\normalfont{f}}}_{\text{approx}}(t)<\uptau^{\,\text{\normalfont{f}}}(t)$.

\begin{figure*}[!ht]%
\centering
\makebox[\textwidth][c]{\includegraphics[width=0.95\textwidth]{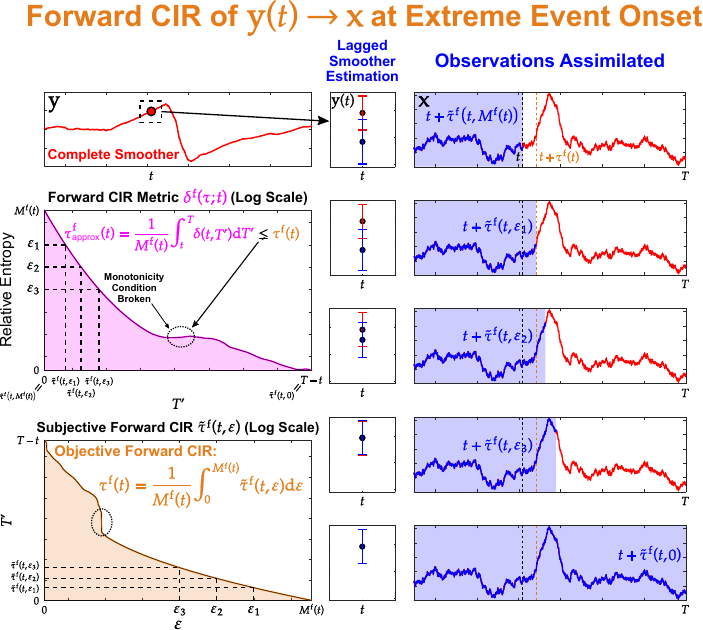}}
\caption{Mathematical details of the forward CIR theory for $\vyt\rightarrow\vx$ at the onset of an extreme event in $\vx$ being generated by $\vyt$. The lagged smoother estimation plots for $\vyt$ (center) showcase how gradual incorporation of future observations of $\vx$ in $(t,T]$ (right) reduces the information deficit from the smoother PDF of $\vyt$ (uncertainty and bias reduction; see Remark \ref{rmk:forward_CIR_properties}), which is quantified by the forward CIR metric in \eqref{eq:forward_CIR_metric} that is plotted on the left for $T'\in[0,T-t]$ for this $t$. On the left the subjective forward CIR length in \eqref{eq:subjective_forward_CIR_length} is also shown for $\varepsilon\in[0,M^{\text{\normalfont{f}}}(t)]$ at this $t$, with the area under its curve (orange), after being normalized by $M^{\text{\normalfont{f}}}(t)$, corresponding to the objective forward CIR length in \eqref{eq:objective_forward_CIR_length}. Since the monotonicity condition of the forward CIR metric is violated for some $T'\in[t,T]$ (black dashed oval), the area under its curved (purple), which corresponds to the computationally efficient approximation of the objective forward CIR length in \eqref{eq:forward_CIR_approximation} after a similar normalization, is a strict underestimate of the exact objective forward CIR.}
\label{fig:details_FCIR}
\end{figure*}

\section{Mathematical Details and Schematic Illustration of the Backward CIR Framework} \label{sec:Backward_CIR_Details}

We start this appendix by presenting a series of remarks outlining the mathematical properties and operational details concerning the (complete) backward CIR metric and its associated subjective and objective lengths.

\begin{remarkapp}[Measurability of the Backward CIR Metric]  \label{rmk:back_CIR_measurability}
    Consider the backward CIR metric from \eqref{eq:backward_CIR_metric}:
    \begin{equation*}
         \delta^{\text{\normalfont{b}}}(\tau;T')=|\delta(\tau T', T')-\delta(0,T')|,
    \end{equation*}
    for $\tau\in\mathrm{I}=[0,1]$ and arbitrary $T>0$ and $T'\in[0,T]$ (of interest is when $T'$ is close to $T$, i.e., in the limit of complete information incorporation). Then, as a continuous affine transformation of the composition between a weakly LSC function (relative entropy) and weakly continuous functions (backward interpolation PDFs and $g(\tau)=\tau T'$), $\delta^{\text{\normalfont{b}}}(\tau;T')$ is a weakly LSC function on $\mathrm{I}$ for each $T>0$ and $T'\in[0,T]$. This weak lower semi-continuity implies that $\delta^{\text{\normalfont{b}}}(\tau;T')$ is also weakly measurable \cite{diestel1977vector, hytnen2016analysis}. One way to establish strong measurability now is to prove that it is also almost surely separably valued as to apply Pettis' measurability theorem \cite{pettis1938integration}, however this step is unnecessary in our setting. For separable spaces, weak and strong measurability coincide, and the space of Borel probability measures defined over a Polish state space is itself Polish, as established by Prokhorov's theorem \cite{van2003probability, kechris1995classical, parthasarathy2005probability}. As such, the backward CIR metric defined in \eqref{eq:backward_CIR_metric} is a measurable function of $\tau \in \mathrm{I}$ for each $T>0$ and $T'\in[0,T]$. Therefore, as the pointwise limit of measurable functions, so is the backward CIR metric after the limit of complete information incorporation, i.e., the complete backward CIR metric $\lim_{T'\to T^-} \delta^{\text{\normalfont{b}}}(\tau;T')$ is also a measurable function of $\tau \in \mathrm{I}$ for each $T>0$.
\end{remarkapp}

\begin{remarkapp}[Temporal Profile and Interpretations of the (Complete) Backward CIR Metric] \label{rmk:backward_CIR_properties}
The key temporal properties of the (complete) backward CIR metric are:
\begin{itemize}
    \item At $\tau=1$, $\delta(T', T')=\mathcal{P}(p^{\text{\normalfont{s}}}_{T'}(\vy|\vx),p_{T'}^{\text{\normalfont{f}}}(\vy|\vx))$ equals the ACI metric assessing $\vy(T')\rightarrow\vx$ (see \eqref{eq:RE_filter_smoother}--\eqref{eq:ACI_cause_notation}), which is then offset by the bias or baseline term $\delta(0, T')$ in \eqref{eq:backward_CIR_metric}, which represents the initial-time information deficit from excluding the data in $(T',T]$ when estimating $\vy(0)$. Typically, in chaotic systems, $M^{\text{\normalfont{b}}}(T)$ occurs at or near $\tau=1$, indicating that effects are primarily attributed to recent causes.

    \item For $\tau\in(0,1)$, $\lim_{T'\to T^-}\delta^{\text{\normalfont{b}}}(\tau;T')$ measures the absolute difference in information gain from incorporating nearly all observations of $\vx$, $T'\to T^-$, when estimating $\vyt=\vy(g(\tau))=\vy(\tau T')$ versus $\vy(0)$. Therefore, the subinterval in $\mathrm{I}$ over which the complete backward CIR metric is large indicates significant causal attribution into the past, as substantial past information gain is ascribed to the currently observed effect $\vx(T)$ beyond the initial-time information deficit baseline.

    \item At $\tau=0$, $\delta^{\text{\normalfont{b}}}(0;T')=0$ by construction.
\end{itemize}
\end{remarkapp}

\begin{remarkapp}[An Inequality for the Subjective Backward CIR Length] \label{rmk:subj_back_CIR_lebesgue}
The bias term $\delta(0,T')$ ensures the supremum in \eqref{eq:subjective_backward_CIR_length} exists, since $0$ always belongs to the set. Furthermore, this also ensures that the subjective backward CIR length underestimates the measurable causal attribution of $\vx(T)$ to $\vy$'s past states in $[0,T]$, since we have that:
\begin{equation*}
    \overset{\sim}{\uptau}^{\,\text{\normalfont{b}}}(T,\varepsilon)\leq T\Big(1-\lambda_{\mathrm{I}}\Big(\Big\{\tau\in\mathrm{I} : \lim_{T'\to T^-}\delta^{\text{\normalfont{b}}}(\tau;T') \leq \varepsilon\Big\}\Big)\Big),
\end{equation*}
with $\lambda_{\mathrm{I}}$ denoting the Lebesgue measure on $\mathrm{I}$.
\end{remarkapp}

\begin{remarkapp}[Operational Details of the Objective Backward CIR] \label{rmk:caveat_back_CIR}
When $M^{\text{\normalfont{b}}}(T)\approx 0$ (weak evidence for $\vy\rightarrow\vx(T)$ assimilative causality), $\uptau^{\,\text{\normalfont{b}}}(T)$ and its computationally efficient (upper-bound) approximation, $\uptau^{\,\text{\normalfont{b}}}_{\text{approx}}(T)$, may yield an inflated CIR length if the complete backward CIR metric gradually increases toward its near-zero maximum $M^{\text{\normalfont{b}}}(T)$ as $\tau$ increases toward the neighborhood of $\tau=1$. Like in Remark~\ref{rmk:caveat_forw_CIR}, such values typically coincide with small or near-zero ACI local maxima and should therefore be construed together with the associated ACI metric as to avoid misidentification of temporal causal attribution. Numerically, to avoid such operationally-inflated values due to floating-point imprecision, if $M^{\text{\normalfont{b}}}(T)<\zeta$ for some small threshold $\zeta>0$ (e.g., $\zeta=10^{-5}$), we can set $\uptau^{\,\text{\normalfont{b}}}(T)=\uptau^{\,\text{\normalfont{b}}}_{\text{approx}}(T)=0$, which will necessarily coincide with $\overset{\sim}{\uptau}^{\,\text{\normalfont{b}}}(T,\varepsilon)\approx0$ for $\varepsilon\geq 0$ by the definition in \eqref{eq:subjective_backward_CIR_length} since $M^{\text{\normalfont{b}}}(T)\approx 0$.
\end{remarkapp}

Figure~\ref{fig:details_BCIR} illustrates the underlying mechanisms of the ACI-based backward CIR theory developed in this work through an illustrative extreme-event scenario (specifically, this analysis corresponds to the example simulation depicted in Panel (II) of Figure~\ref{fig:schematic_diagram}). We focus on a time $T$ where an extreme event in $\vx$ caused by $\vy$ reaches its peak amplitude $\vx(T)$. Both the objective and subjective backward CIR formulations are depicted. To understand the results shown, note that the subjective backward CIR length in \eqref{eq:subjective_backward_CIR_length} is an affine transformation of the generalized inverse of the complete backward CIR metric in \eqref{eq:complete_backward_CIR_metric} \cite{feng2012note, embrechts2013note, kampke2014generalized}. For this observational time $T$ and observed effect $\vx(T)$, the complete backward CIR metric is an increasing and continuous function of natural time $t$. Hence, unlike the results in Figure~\ref{fig:details_FCIR}, a bijection exists between the subjective backward CIR length and the inverse of the complete backward CIR metric, while this confirms that the computationally efficient approximation in \eqref{eq:backward_CIR_metric} precisely recovers the associated objective CIR, $\uptau^{\,\text{\normalfont{b}}}_{\text{approx}}(T)=\uptau^{\,\text{\normalfont{b}}}(T)$, since the monotonicity condition in Theorem~\ref{thm:obj_backward_CIR_approx} is satisfied.

\begin{figure*}[!ht]%
\centering
\makebox[\textwidth][c]{\includegraphics[width=0.95\textwidth]{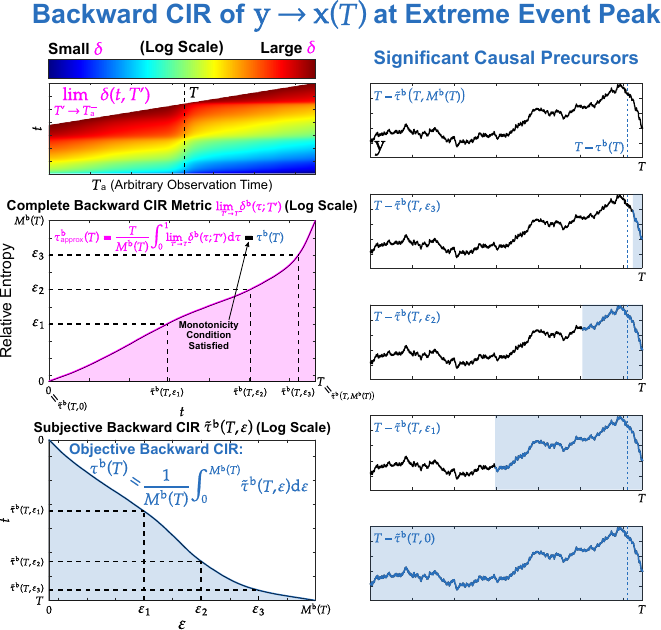}}
\caption{Mathematical details of the backward CIR theory for $\vy\rightarrow\vx(T)$ at the peak of an extreme event $\vx(T)$ causally produced by $\vy$. On the left, the complete backward CIR metric in \eqref{eq:backward_CIR_metric}--\eqref{eq:complete_backward_CIR_metric} is plotted for $t\in[0,T]$ and arbitrary observational times and at this $T$, alongside the subjective backward CIR length in \eqref{eq:subjective_backward_CIR_length} for $\varepsilon\in[0,M^{\text{\normalfont{b}}}(T)]$ at this $T$. The former quantifies the decay in information deficit in the lagged smoother PDF for $\vyt$ from using $\vx(s\leq T-\dt)$, for $\dt\ll 1$, from the complete smoother PDF with $\vx(s\leq T)$, beyond the initial-time baseline as $t$ decreases from $T$ to $0$ (see Remark \ref{rmk:backward_CIR_properties}), illustrated by the $\vy$ plots on the right for various thresholds $\varepsilon$ and corresponding subjective backward CIRs. Since the monotonicity condition of the complete backward CIR metric is satisfied for all $t\in[0,T]$, the areas under the complete backward CIR metric's curve (purple) and that of the subjective backward CIR length (blue) are equal, which, after being normalized by $M^{\text{\normalfont{b}}}(T)$, correspond to the computationally efficient approximation (see Eq.~\eqref{eq:obj_backward_CIR_approx}) and definition (see Eq.~\eqref{eq:objective_backward_CIR_length}) of the objective backward CIR length, respectively.}
\label{fig:details_BCIR}
\end{figure*}

\section{Physical and Operational Interpretations of the Forward and Backward CIRs and Their Relationship} \label{sec:CIR_Connections}

In this section, we provide some intuitive interpretations of the forward and backward CIRs and their mathematical relationship, so as to clarify their physical, practical, and operational significance for quantifying the temporal extent of causality in complex dynamical systems.

For concrete operational analogs of the subjective and objective CIRs, both forward and backward, we can look into correlation analysis. Specifically, the subjective and objective CIRs can be regarded as information-theoretic causal analogs of the autocorrelation function and the decorrelation time, respectively, which are statistical metrics used to measure the memory of a stochastic system \cite{andreou2026assimilative}. Likewise to the subjective CIRs, the auto- and cross-correlation functions are often utilized to assess the temporal persistence of stochastic processes by measuring the duration of time over which the correlation remains above a certain threshold. On the other hand, similar to the objective CIR, the decorrelation time integrates the associated correlation function over time to quantify the effective memory length of the process without relying on empirical thresholds. 

Going further into their mathematical relationship, intuitively, the forward and backward CIRs provide complementary temporal perspectives and a time dimension to state-space causality, assigning temporal extents in opposite directions:
\begin{enumerate}[label=\textbf{(\Alph*)}]
    \item Forward CIR: Characterizes the {future} persistence of $\vyt$'s causal influence on $\vx$, contingent on evidence for $\vyt\rightarrow \vx$ at time $t\in[0,T]$.
    \item Backward CIR: Locates the {past} onset of causal precursors $\vy$ for the observed effect $\vx(T)$, conditional on significant evidence toward ${\vy\rightarrow\vx(T)}$ at time $T>0$.
\end{enumerate}
This complementary nature between the two CIRs is further substantiated by the evident mathematical duality between their parallel properties that are developed and discussed in Sections~\ref{sec:Forward_CIR} and~\ref{sec:Backward_CIR} and in Appendix~\ref{sec:Forward_CIR_Details}--\ref{sec:Backward_CIR_Details}. Compare for the forward CIR with the backward one, respectively: Eq.~\eqref{eq:forward_CIR_approximation} (lower-bound/``$\leq$'' approximation) with Eq.~\eqref{eq:obj_backward_CIR_approx} (upper-bound/``$\geq$'' approximation) and when respective equality is achieved (nonincreasing forward CIR metric vs nondecreasing complete backward CIR metric), Remark~\ref{rmk:approx_obj_forw_CIR_interval}  with Remark~\ref{rmk:approx_obj_back_CIR_interval}, Remarks~\ref{rmk:forward_CIR_properties}--\ref{rmk:caveat_forw_CIR} with Remarks~\ref{rmk:backward_CIR_properties}--\ref{rmk:caveat_back_CIR} (dual temporal profiles for the forward and complete backward CIR metrics and overestimation of causal prediction vs underestimation of causal attribution), and Figure~\ref{fig:details_FCIR} with Figure~\ref{fig:details_BCIR}. Together, these mathematically rigorous notions of forward and backward CIRs establish a unified and comprehensive theory for the quantification of causal prediction and attribution in complex dynamical systems within the ACI framework.

Finally, their interpretations and relationship naturally mirrors fundamental concepts from physics. In the relativistic space-time continuum, the forward CIR defines the {future light cone} of $\vyt$ in the $\vx$-marginal state space, while the backward CIR defines the {past light cone} of $\vx(T)$ in the $\vy$-marginal state space \cite{sard1970relativistic, penrose1972techniques}. More concretely, hyperbolic PDEs can also provide a direct analogy through their finite information propagation rate \cite{evans2010partial}. The forward CIR mirrors the {range of influence}, delimiting how $\vyt$'s effects propagate in the future, while the backward CIR corresponds to the {domain of dependence}, identifying which past $\vy$ values caused $\vx(T)$. However, unlike fixed characteristic cones, the CIRs of complex systems are dynamical regions that evolve with the causal flow between the $\vy$ and $\vx$ subspaces.

\section{Proof of Theorem~\ref{thm:obj_backward_CIR_approx}} \label{sec:Comp_Back_CIR_Thm_Proof}

Here we provide the proof of Theorem~\ref{thm:obj_backward_CIR_approx}.

\begin{proof}
   First, observe that $\lambda_\mathrm{I}(\boldsymbol{\cdot})$ defines a probability measure on $\mathrm{I} = [0,1]$. Under the (Lebesgue) measurability assumption for $\lim_{T'\to T^-}\delta^{\text{\normalfont{b}}}(\boldsymbol{\cdot};T')$, we can then interpret it as a continuous random variable on the probability space $(\mathrm{I}, \mathcal{L}_\mathrm{I}, \lambda_\mathrm{I})$ with its support being $[0, M^{\text{\normalfont{b}}}(T)]$, where $\mathcal{L}_\mathrm{I}$ is the $\sigma$-algebra of Lebesgue measurable subsets of $\mathrm{I}$.

   Its survival function,
   \begin{equation*}
       1-\lambda_{\mathrm{I}}\Big(\Big\{\tau\in\mathrm{I} : \lim_{T'\to T^-}\delta^{\text{\normalfont{b}}}(\tau;T') \leq \varepsilon\Big\}\Big),
   \end{equation*}
   is well-defined because (a) it is nonincreasing and right-continuous with left limits, (b) it equals $1$ for $\varepsilon \leq 0$, and (c) it equals zero for $\varepsilon \geq M^{\text{\normalfont{b}}}(T)$.

   Applying the expected survival time formula to this nonnegative, compactly supported random variable yields:
    \begin{align*}
        \mathbb{E}\Big[\lim_{T'\to T^-}\delta^{\text{\normalfont{b}}}(\tau;T')\Big]
        &:= \int_0^1 \lim_{T'\to T^-}\delta^{\text{\normalfont{b}}}(\tau;T') \rmd\tau\\
        &= \int_0^{M^{\text{\normalfont{b}}}(T)} \Big(1-\lambda_{\mathrm{I}}\Big(\Big\{\tau\in\mathrm{I} : \lim_{T'\to T^-}\delta^{\text{\normalfont{b}}}(\tau;T') \leq \varepsilon\Big\}\Big)\Big)\rmd\varepsilon.
    \end{align*}

    The theorem's main result then follows by dividing through by $M^{\text{\normalfont{b}}}(T) > 0$, multiplying by $T$, and applying the inequality from Remark~\ref{rmk:subj_back_CIR_lebesgue}:
    \begin{equation*}
        \overset{\sim}{\uptau}^{\,\text{\normalfont{b}}}(T,\varepsilon)\leq T\Big(1-\lambda_{\mathrm{I}}\Big(\Big\{\tau\in\mathrm{I} : \lim_{T'\to T^-}\delta^{\text{\normalfont{b}}}(\tau;T') \leq \varepsilon\Big\}\Big)\Big).
    \end{equation*}
    The equality condition also follows immediately from this, since equality in the above holds if and only if $\lim_{T'\to T^-}\delta^{\text{\normalfont{b}}}(\boldsymbol{\cdot};T')$ is nondecreasing in $\mathrm{I}$.
\end{proof}

\section{Conditionally Linear Dynamics with State-Independent Feedbacks as Representatives of Complex Systems} \label{sec:Linear_Dynamics_y}

While the system \eqref{eq:reduced_model} in Section~\ref{sec:Backward_CIR_Linear} might appear overly simplistic so as to obtain the result in Theorem~\ref{thm:obj_backward_CIR_linear}, many complex stochastic nonlinear dynamical systems with multivariate components $\vx$ and $\vy$ can be put in such a reduced-order canonical form. In this section, we provide a brief overview of how this is possible through some practical generalizations:
\begin{itemize}
    \item In \eqref{eq:reduced_model}, the noise feedbacks are purely additive or state independent. This is without loss of generality, as transitioning from state-dependent noise feedbacks to fully additive ones is possible via the use of a suitable Lamperti transform (which adjusts the system's potential to account for the transition from multiplicative to additive noise) \cite{moller2010state}.

    \item In \eqref{eq:reduced_model}, noise cross-interactions are assumed to be nonexistent, i.e., $W_1$ does not appear in $y$'s evolution nor does $W_2$ in $x$'s equation. The case of correlated noise forcings can be easily reduced to the uncorrelated one via a whitening transformation due to the linear mean dynamics and state-independent noise forcings \cite{rozovsky2012stochastic, liptser2001statistics}.
    \begin{itemize}[label=$\blacktriangleright$]
        \item We note here that the proof of Theorem~\ref{thm:obj_backward_CIR_linear} (see Appendix~\ref{sec:Back_CIR_Linear_Thm_Proof}) can be adapted to provide an analogous result in the case where there are cross-interacting or colored noise feedbacks, without the need for whitening, by accounting for their correlations in the auxiliary components of the online smoother formulae (see Theorem~\ref{thm:onlinesmoother}) \cite{andreou2026adaptive}.
    \end{itemize}

    \item In \eqref{eq:reduced_model}, we assume $x$ and $y$ are one-dimensional. This is without loss of generality, as it is known that a diagonal approximation of the filter and smoother posterior covariance matrices is possible for suitable multidimensional variants \cite{chen2014information}.

    \item In \eqref{eq:reduced_model}, $y$'s feedbacks, both in $x$ and in $y$, are constant in time. For time-varying coefficients (possibly also depending on $x$), which satisfy sufficient conditions for well-posedness, stochastic linear stability, and the existence of equilibrium filter and smoother PDFs (posterior statistical attractors), then a simple linearization procedure can be utilized to retrieve similar results to those established in Theorem~\ref{thm:obj_backward_CIR_linear}.
\end{itemize}

\section{Computational Tool for Exact CIR Inference: Conditional Gaussian Nonlinear Systems} \label{sec:Tool_CGNS}

Computing objective CIR lengths requires evaluating the CIR metrics $\delta^{\text{\normalfont{f}}}(\tau;t)$ and $\lim_{T'\to T^-}\delta^{\text{\normalfont{b}}}(\tau;T')$, which depend on the lagged and complete smoother PDFs. While for the general nonlinear system in \eqref{eq:general_CTNDS} ensemble methods have to be utilized \cite{jiang2026continuous}, a wide range of nonlinear systems, called conditional Gaussian nonlinear systems (CGNSs), admit closed-form solutions for both filter and smoother posterior distributions. These systems take the form  \cite{liptser2001statistics, chen2018conditional, chen2022conditional}:\begin{subequations} \label{eq:CGNS}
    \begin{align}
        \frac{\rmd\vxt}{\rmd t} &= \big(\ml^\vx(t,\vx)\vyt+\vf^\vx(t,\vx)\big)+\ms_1^\vx(t,\vx) \dot{\vw}_1(t)+\ms_2^\vx(t,\vx)\dot{\vw}_2(t), \label{eq:CGNS1}\\
        \frac{\rmd\vyt}{\rmd t} &= \big(\ml^\vy(t,\vx)\vyt+\vf^\vy(t,\vx)\big)+\ms_1^\vy(t,\vx)\dot{\vw}_1(t)+\ms_2^\vy(t,\vx)\dot{\vw}_2(t), \label{eq:CGNS2}
    \end{align}
\end{subequations}
where $\vx$ appears nonlinearly while $\vy$ exclusively does so linearly. Nevertheless, the marginal distributions of both $\vx$ and $\vy$, as well as their joint distribution, can be highly non-Gaussian. Only the conditional distributions of $\vy$ given a realization of $\vx$ are Gaussians, including the posterior filter and smoother \cite{liptser2001statistics}. CGNSs are widely applicable in neuroscience, ecology, and geophysics due to their efficacy in continuous data assimilation.

We employ an online smoother for CGNSs \cite{andreou2026adaptive} that enables exact computation of both CIR metrics. This algorithm efficiently computes lagged smoother PDFs in real-time by adaptively performing only the most influential updates due to a new observation, thus capturing the essential temporal information for state estimation when assimilating the new data while minimizing computational costs. Appendix~\ref{sec:Online_Smoother}--\ref{sec:Tool_Backward_CIR} outline this methodology and its application in computing the forward and backward CIR metrics and lengths.

\subsection{An Online Smoother for CGNSs and its Use in Deriving Closed-Form Expressions for the CIR Lengths} \label{sec:Online_Smoother}

In this section, we briefly review the pertinent results from \cite{andreou2026adaptive}, which developed an online smoother for CGNSs. In the subsequent subsections, we outline how the main theorem cited here can be applied to compute the complete and lagged smoother PDFs from Section~\ref{sec:CIR_Framework} in the case of a CGNS. These distributions are required for computing the forward and (complete) backward CIR metrics from Sections~\ref{sec:Forward_CIR} and~\ref{sec:Backward_CIR}, which, in turn, are needed for evaluating the corresponding lower- and upper-bound approximations of the objective forward and backward CIR lengths, respectively.

We first adopt a uniform discretization of the interval $[0,T]$, transitioning to discrete time, with $0 = t_0 < t_1 < \cdots < t_N = T$ and $t_j = j\Delta t$, where $j \in \llbracket N\rrbracket := \{0,1,\ldots,N\}$ and $N \in \mathbb{N}$, with $N$ large enough so that the partition norm $\Delta t = T/N$ satisfies $0 < \Delta t \ll 1$. In what follows, $T$ is assumed to be arbitrary and allowed to increase freely in $(0,+\infty)$, modeling the successive observation of $\vx$, with $\dt\ll 1$ predetermined from either the model simulation's integration time-step or observation rate. (The assumption of uniform $\dt$ is without loss of generality \cite{andreou2026adaptive}.) Therefore, $N:=\lceil T/\dt\rceil$ with $T$ being sequentially incremented by $\dt$. Under this time discretization, both the natural ($t$) and observational ($T'$ and $T$) time variables of $\delta$ in \eqref{eq:metric_CIR} take values in the discrete set $\{t_j\}_{j \in \llbracket N \rrbracket}$. We use the index $j$ to denote the discretization along natural time and $n$ over observational time, while the superscript notation $\boldsymbol{\cdot}^{\,j}$ is used to denote the discrete-time approximation of the corresponding continuous vector- or matrix-valued function evaluated at $t_j$; for example, $\vx^j := \vx(t_j)$ and $\ml^{\vx,j} := \ml^\vx(t_j, \vx(t_j))$. Additionally, we write $\big(\boldsymbol{\cdot} \big| \vx(s \leq n)\big)$ to indicate that we condition on the observations $\{\vx^j\}_{j \in \llbracket n \rrbracket}$ for $n \in \llbracket N \rrbracket$. For arbitrary $j$, we define the following auxiliary matrices:
\begin{align*}
    \mathbf{E}^j&:=\mathbf{I}_{l\times l}+\left[(\ms^\vy\circ\ms^\vx)^j\big((\ms^\vx\circ\ms^\vx)^{j}\big)^{-1}\mathbf{G}^{\vx,j}-\mathbf{G}^{\vy,j}\right]\dt\in\mathbb{R}^{l\times l},\\
    \begin{split}
        \mathbf{F}^j&:=-\mr{\nf}^j\Big[(\mathbf{K}^j)^{\mathtt{T}}+\left((\mathbf{G}^{\vx,j})^\mathtt{T}\mathbf{K}^j\mr{\nf}^j(\mathbf{K}^j)^{\mathtt{T}}-(\mr{\nf}^j)^{-1}(\mathbf{H}^{j})^\mathtt{T}\mr{\nf}^j(\mathbf{K}^j)^\mathtt{T}+(\ml^{\vy,j})^{\mathtt{T}}(\mathbf{K}^{j})^\mathtt{T}\right)\dt\\
        &\hspace*{5cm} -(\ml^{\vx,j})^\mathtt{T}\left(\big((\ms^\vx\circ\ms^\vx)^{j}\big)^{-1}+\mathbf{K}^j\mr{\nf}^j(\mathbf{K}^j)^\mathtt{T}\dt\right)\Big]\in\mathbb{R}^{l\times k},
    \end{split}
\end{align*}
where the total row-based Gramians encode the noise-feedback/diffusion-coefficient interactions and are defined as
\begin{equation*}
    (\ms^{\text{\textbullet}}\circ\ms^{\scalebox{0.4}{\(\blacksquare\)}})^j:=\ms_1^{\text{\textbullet},j}(\ms_1^{\scalebox{0.4}{\(\blacksquare\)},j})^\mathtt{T}+\ms_2^{\text{\textbullet},j}(\ms_2^{\scalebox{0.4}{\(\blacksquare\)},j})^\mathtt{T}\in\mathbb{R}^{\mathrm{dim}(\text{\textbullet})\times \mathrm{dim}(\scalebox{0.4}{\(\blacksquare\)})}, \qquad \text{\textbullet}, \,\scalebox{0.6}{\(\blacksquare\)}\in\{\vx,\vy\},
\end{equation*}
while
\begin{gather*}
    \mathbf{G}^{\vx,j}:=\ml^{\vx,j}+(\ms^\vx\circ\ms^\vy)^j(\mr{\nf}^j)^{-1}\in\mathbb{R}^{k\times l}, \quad \mathbf{G}^{\vy,j}:=\ml^{\vy,j}+(\ms^\vy\circ\ms^\vy)^j(\mr{\nf}^j)^{-1}\in\mathbb{R}^{l\times l}, \\
    \mathbf{H}^j:=(\mr{\nf})^{-1}\left(\ml^{\vy,j}\mr{f}^j+\mr{f}^j(\ml^{\vy,j})^{\mathtt{T}}+(\ms^\vy\circ\ms^\vy)^j\right)\in\mathbb{R}^{l\times l},\\
    \mathbf{K}^j:=\big((\ms^\vx\circ\ms^\vx)^{j}\big)^{-1}\mathbf{G}^{\vx,j}\in\mathbb{R}^{k\times l}.
\end{gather*}

Under the preceding exposition, the following theorem holds, characterizing the online smoother distributions of the CGNS framework in discrete-time settings \cite{andreou2026adaptive}. Specifically, it outlines the procedure for online estimation of the smoother PDF in a CGNS, i.e., its iterative calculation at each natural time instant when considering a newly obtained observation from $\vx$.

\begin{theorem}[Optimal Online Forward-In-Time Discrete Smoother] \label{thm:onlinesmoother}
Let $\vxt$ and $\vyt$ satisfy \eqref{eq:CGNS1}--\eqref{eq:CGNS2}. Then, under suitable regularity conditions, the discrete smoother (backward interpolation) distributions are Gaussian,
\begin{equation*}
    \pp\big(\vy^j\big|\vx(s\leq n)\big) \overset{(\rmd)}{\sim}\mathcal{N}_l(\vm{s}^{j,n},\mr{s}^{j,n}),
\end{equation*}
for $j\in\llbracket n\rrbracket$, $n\in\llbracket N\rrbracket$, where
\begin{gather*}
    \vm{\normalfont{s}}^{j,n}:=\ee{\vy^j\big|\vx(s\leq n)},\quad \mr{\normalfont{s}}^{j,n}:=\ee{(\vy^j-\vm{\normalfont{s}}^{j,n})(\vy^j-\vm{\normalfont{s}}^{j,n})^\mathtt{T}\big|\vx(s\leq n)},
\end{gather*}
satisfy the following recursive backward difference equations for $j\in\llbracket n-2\rrbracket$ and $n\in\{2,\ldots,N\}$:
\begin{subequations} \label{eq:onlinerecursive}
    \begin{align}
        \vm{\ns}^{j,n}&=\vm{\ns}^{j,n-1}+\mathbf{D}^{j,n-2}\left(\vm{\ns}^{n-1,n}-\vm{\nf}^{n-1}\right), \label{eq:onlinerecursive1}\\
        \mr{\ns}^{j,n}&=\mr{\ns}^{j,n-1}+\mathbf{D}^{j,n-2}\left(\mr{\ns}^{n-1,n}-\mr{\nf}^{n-1}\right)(\mathbf{D}^{j,n-2})^\mathtt{T},\label{eq:onlinerecursive2}
    \end{align}
\end{subequations}
where the update matrices $\mathbf{D}^{j,n-2}$ are defined by
\begin{equation} \label{eq:updatematrix}
    \mathbf{D}^{n-1,n-2}:=\mathbf{I}_{l\times l} \quad\&\quad \mathbf{D}^{j,n-2}:=\overset{\mathlarger{\curvearrowright}}{\prod^{n-2}_{i=j}} \mathbf{E}^i:=\mathbf{E}^j\mathbf{E}^{j+1}\cdots \mathbf{E}^{n-2},
\end{equation}
while for $j=n-1$ and $n\in\{1,\ldots,N\}$ we have:
\begin{align*}
    \vm{\ns}^{n-1,n}&=\mathbf{E}^{n-1}\vm{\nf}^{n}+\mathbf{b}^{n-1}, \\
    \mr{\ns}^{n-1,n}&=\mathbf{E}^{n-1}\mr{\nf}^{n}(\mathbf{E}^{n-1})^\mathtt{T}+\mathbf{P}^{n-1}_n,
\end{align*}
with the $\mathbf{b}^{n-1}$ and $\mathbf{P}^{n-1}_n$ auxiliary residual terms being given as
\begin{align*}
    \mathbf{b}^{n-1}&:=\vm{\nf}^{n-1}-\mathbf{E}^{n-1}\left((\mathbf{I}_{l\times l}+\ml^{\vy,n-1}\dt)\vm{\nf}^{n-1}+\vf^{\vy,n-1}\dt\right)\\
    &\hspace{1.55cm}+\mathbf{F}^{n-1}\left(\vx^{n}-\vx^{n-1}-(\ml^{\vx,n-1}\vm{\nf}^{n-1}+\vf^{\vx,n-1})\dt\right),\\
    \mathbf{P}^{n-1}_{n}&:= \mr{\nf}^{n-1}-\mathbf{E}^{n-1}(\mathbf{I}_{l\times l}+\ml^{\vy,n-1}\dt)\mr{\nf}^{n-1}-\mathbf{F}^{n-1}\ml^{\vx,n-1}\mr{\nf}^{n-1}\dt,
\end{align*}
and finally for $j=n$ and $n\in\llbracket N\rrbracket$ we have $\vm{\ns}^{n,n}=\vm{\nf}^n$ and $\mr{\ns}^{n,n}=\mr{\nf}^n$, since the smoother and filter posterior Gaussian statistics coincide at the right endpoint by definition.
\end{theorem}
The Gaussianity of the discrete-time posterior distributions of a CGNS is proved in Chapter 13 of Liptser \& Shiryaev \cite{liptser2001statistics} (see also Lemma A6 in \cite{andreou2024martingale} for the discrete-time optimal nonlinear smoother distribution proof specifically), while the proof of Theorem~\ref{thm:onlinesmoother} is given in Theorem 3.1 of \cite{andreou2026adaptive}.

In the sequel, we present the details for explicitly computing the approximations of the objective forward and backward CIR lengths in the unconditional setting (as given in \eqref{eq:forward_CIR_approximation} and \eqref{eq:obj_backward_CIR_approx}, respectively) via analytical means. The same methodology extends naturally to conditional assimilative causal links, with the appropriate modifications. As for computing the (conditional) ACI metric in \eqref{eq:relative_entropy}--\eqref{eq:RE_filter_smoother} (\eqref{eq:RE_filter_smoother_general}--\eqref{eq:filter_smoother_ancillary_inf_uncert}) for CGNSs, the details are given in \cite{andreou2026assimilative}.

\subsection{Computation of the Forward CIR in CGNSs} \label{sec:Tool_Forward_CIR}

From Theorem~\ref{thm:onlinesmoother}, by adopting the following notation for $j\in\llbracket N\rrbracket$, \linebreak $n\in\{j,j+1,\ldots,N\}$:
\begin{equation} \label{eq:updatedlaggeddistr_CIR}
    p_n(\vy^j|\vx) \text{ corresponding to } \mathcal{N}_l(\vm{s}^{j,n},\mr{s}^{j,n}),
\end{equation}
we have for
\begin{equation*}
    \mathcal{P}^{j}_n:=\mathcal{P}\big(p_N(\vy^j|\vx), p_n(\vy^j|\vx)\big),
\end{equation*}
that:
\begin{align}
\begin{split}
    \mathcal{P}^{j}_n=&\ \frac{1}{2}\left(\vm{s}^{j,N}-\vm{s}^{j,n}\right)^\mathtt{T}(\mr{s}^{j,n})^{-1}\left(\vm{s}^{j,N}-\vm{s}^{j,n}\right)\\
    &+\frac{1}{2}\left(\mathrm{tr}\big(\mr{s}^{j,N}(\mr{s}^{j,n})^{-1}\big)-l-\log\big(\mathrm{det}\big(\mr{s}^{j,N}(\mr{s}^{j,n})^{-1}\big)\big)\right),
\end{split} \label{eq:lackinfoupdate_CIR}
\end{align}
due to the {signal--dispersion decomposition} of the relative entropy for Gaussian PDFs \cite{andreou2026assimilative, cai2002mathematical}: The quadratic-form term in \eqref{eq:lackinfoupdate_CIR} is the {signal} part, while the term equal to half the Burg or log-det divergence from $\mr{s}^{j,N}$ to $\mr{s}^{j,n}$, is the {dispersion} part. Then, by the theory in Section~\ref{sec:Forward_CIR}, we have from \eqref{eq:forward_CIR_metric} the following exact equality:
\begin{equation} \label{eq:approx_info_gains_cgns_CIR}
    \delta(t_j,t_n)=\mathcal{P}^{j}_n, \quad j\in\llbracket N\rrbracket,\ n\in\{j,j+1,\ldots,N\}.
\end{equation}
As such, in the case of a CGNS, we have that the discrete-time subjective forward CIR length of $\vy(t_j)\rightarrow\vx$ for the threshold parameter $\varepsilon$ is given by (see \eqref{eq:subjective_forward_CIR_length}):
\begin{equation} \label{eq:subjective_CIR_length_cgns}
    \overset{\sim}{\uptau}^{\,\text{\normalfont{f}}}(t_j,\varepsilon)= \underset{n\in\{j,\ldots,N\}}{\max}\big\{\mathcal{P}^{j}_n> \varepsilon\big\}\dt-t_j\in[0,T-t_j],
\end{equation}
which subsequently defines its objective discrete-time counterpart via \eqref{eq:objective_forward_CIR_length}, by averaging \eqref{eq:subjective_CIR_length_cgns} over $\varepsilon$. In addition, we have that the lower-bound approximation of the associated objective forward CIR length in the case of discrete time for a CGNS, is given by (see \eqref{eq:forward_CIR_approximation}):
\begin{equation} \label{eq:objective_CIR_length_cgns}
    [0,T-t_j]\ni\frac{\big\lVert \mathcal{P}^j_{\boldsymbol{\cdot}}\big\rVert_{L^1(\{j,\ldots,N\})}}{\big\lVert \mathcal{P}^j_{\boldsymbol{\cdot}}\big\rVert_{L^\infty(\{j,\ldots,N\})}}=\uptau^{\,\text{\normalfont{f}}}_{\text{approx}}(t_j)\leq \uptau^{\,\text{\normalfont{f}}}(t_j),
\end{equation}
where for $j\in\llbracket N\rrbracket$ and a bounded grid function $g_n=g(t_n)$ defined on the equidistant grid points $\{t_n\}_{n\in\{j,j+1,\ldots,N\}}$ (uniform partition of $[0,T]$ with mesh size of $\dt$), its $L^1$ and $L^{\infty}$ norms on $\{j,j+1,\ldots,N\}$ are defined as \cite{leveque2007finite}:
\begin{gather*}
    \lVert g\rVert_{L^1(\{j,\ldots,N\})}=\dt\sum_{n=j}^N |g(t_n)|,\\
    \lVert g\rVert_{L^\infty(\{j,\ldots,N\})}=\underset{n\in\{j,\ldots,N\}}{\max}\{|g(t_n)|\},
\end{gather*}
essentially approximating the analytic Riemann integral with its associated right Riemann sum.

We note that, computationally (as implemented in the paper's codebase), the $p_n(\vy^j|\vx)$ PDFs in \eqref{eq:updatedlaggeddistr_CIR} are obtained by using the so-called fixed-lag variant of the CGNS online smoother in Theorem~\ref{thm:onlinesmoother} \cite{andreou2026adaptive}. Essentially, all of the updates in the recursive equations in \eqref{eq:onlinerecursive} are carried out when transitioning and assimilating a newly obtained observation $\vx^n$ onto ${\vx(s\leq n-1)}$, which we do for all $n\in\llbracket N\rrbracket$. This means a full fixed-lag value is used at each new observation, which translates to carrying out the full forward filter--backward smoother pipeline for each $n$ \cite{andreou2026adaptive}.

\subsection{Computation of the Backward CIR in CGNSs} \label{sec:Tool_Backward_CIR}

From Theorem~\ref{thm:onlinesmoother}, for $\dt$ sufficiently small ($\dt\ll 1$), by adopting the following notation for $j\in\llbracket N\rrbracket$ with $N$ arbitrary like $T$, which are both sequentially incremented (latter by $\dt$) with each new observation $\vx^N$ (see also definitions of complete and lagged smoother PDFs in Section~\ref{sec:CIR_Framework}):
\begin{equation}
    \begin{gathered}
        p_T^{\text{upd}}(\vy^j|\vx) \text{ corresponding to } \mathcal{N}_l(\vm{s}^{j,N},\mr{s}^{j,N}) \ (n=N;\text{ complete smoother}),\\
        p_T^{\text{lag}}(\vy^j|\vx) \text{ corresponding to } \mathcal{N}_l(\vm{s}^{j,N-1},\mr{s}^{j,N-1}) \  (n=N-1;\text{ lagged smoother}),
    \end{gathered} \label{eq:updatedlaggeddistr_OIR}
\end{equation}
where we have used $T'=T-\dt$ for the lagged smoother, we have for
\begin{equation*}
    \mathcal{P}^{j}_T:=\mathcal{P}\big(p_T^{\text{upd}}(\vy^j|\vx), p_T^{\text{lag}}(\vy^j|\vx)\big), \quad T>0,
\end{equation*}
that \cite{andreou2026adaptive}:
\begin{align}
\begin{split}
    \mathcal{P}^{j}_T=&\ \frac{1}{2}\left(\vm{s}^{N-1,N}-\vm{f}^{N-1}\right)^\mathtt{T}(\mathbf{D}^{j,N-2})^\mathtt{T}(\mr{s}^{j,N-1})^{-1}\mathbf{D}^{j,N-2}\left(\vm{s}^{N-1,N}-\vm{f}^{N-1}\right)\\
    &+\frac{1}{2}\left(\mathrm{tr}(\mathbf{Q}^{j,N})-l-\log(\mathrm{det}(\mathbf{Q}^{j,N}))\right),
\end{split} \label{eq:lackinfoupdate_OIR}
\end{align}
due to the signal--dispersion decomposition of the relative entropy for Gaussian PDFs \cite{andreou2026assimilative, cai2002mathematical}, where $\mathbf{Q}^{j,N}$ is defined as the covariance ratio matrix \cite{andreou2026adaptive}:
\begin{align}
\begin{split}
    \mathbf{Q}^{j,N}&:=\left(\mr{s}^{j,N-1}+\mathbf{D}^{j,N-2}\left(\mr{s}^{N-1,N}-\mr{f}^{N-1}\right)(\mathbf{D}^{j,N-2})^\mathtt{T}\right)(\mr{s}^{j,N-1})^{-1}\\
    &=\mathbf{I}_{l\times l}+\left(\mathbf{D}^{j,N-2}\left(\mr{s}^{N-1,N}-\mr{f}^{N-1}\right)(\mathbf{D}^{j,N-2})^\mathtt{T}\right)(\mr{s}^{j,N-1})^{-1}.
\end{split} \label{eq:covarianceratio_OIR}
\end{align}
(The dependence of $\mathcal{P}^{j}_T$ on $T$ is explicitly noted here for clarity.) Then, by the theory in Section~\ref{sec:Backward_CIR}, we have from \eqref{eq:backward_CIR_metric} the following approximation:
\begin{equation} \label{eq:approx_info_gains_cgns_OIR}
    \lim_{T'\to T^-}\delta(t_j,T')\approx\mathcal{P}^{j}_T, \quad j\in\llbracket N\rrbracket.
\end{equation}
Note that this is an approximation, unlike the equality in \eqref{eq:approx_info_gains_cgns_CIR}, as this result is contingent on $t_{N}-t_{N-1}=\dt=T/N\ll 1$ being sufficiently small so that the limit of complete information incorporation, $T'\to T^-$, is sufficiently well approximated in discrete time. As such, in the case of a CGNS, we have that the discrete-time subjective backward CIR length of $\vy\rightarrow\vx(T)$ for the threshold parameter $\varepsilon\geq 0$ can be approximated as (see \eqref{eq:subjective_backward_CIR_length}):
\begin{equation} \label{eq:subjective_OIR_length_cgns}
    \overset{\sim}{\uptau}^{\,\text{\normalfont{b}}}(T,\varepsilon) \approx T-\underset{j\in\llbracket N\rrbracket}{\sup}\big\{\big|\mathcal{P}^{j}_T-\mathcal{P}^{0}_T\big|\leq \varepsilon\big\}\dt\in[0,T],
\end{equation}
which subsequently approximates its objective discrete-time counterpart via \eqref{eq:objective_backward_CIR_length}, by averaging \eqref{eq:subjective_OIR_length_cgns} over $\varepsilon$. Similarly, we have that the upper-bound approximation of the associated objective backward CIR length in the case of discrete time for a CGNS, is in turn approximated as (see \eqref{eq:obj_backward_CIR_approx} and Theorem~\ref{thm:obj_backward_CIR_approx}):
\begin{equation} \label{eq:objective_OIR_length_cgns}
     \uptau^{\,\text{\normalfont{b}}}(T) \leq \uptau^{\,\text{\normalfont{b}}}_{\text{approx}}(T)\approx\frac{\big\lVert \mathcal{P}^{\boldsymbol{\cdot}}_T-\mathcal{P}^{0}_T\big\rVert_{L^1(\llbracket N\rrbracket)}}{\big\lVert \mathcal{P}^{\boldsymbol{\cdot}}_T-\mathcal{P}^{0}_T\big\rVert_{L^\infty(\llbracket N\rrbracket)}}\in[0,T],
\end{equation}
where for $j\in\llbracket N\rrbracket$ and a bounded grid function $g^j=g(t_j)$ defined on the equidistant grid points $\{t_j\}_{j\in\llbracket N\rrbracket}$ (uniform partition of $[0,T]$ with mesh size of $\dt$), its $L^1$ and $L^{\infty}$ norms on $\llbracket N\rrbracket$ are defined as \cite{leveque2007finite}:
\begin{gather*}
    \lVert g\rVert_{L^1(\llbracket N\rrbracket)}=\dt\sum_{j=0}^N |g(t_j)|,\\
    \lVert g\rVert_{L^\infty(\llbracket N\rrbracket)}=\underset{j\in\llbracket N\rrbracket}{\max}\{|g(t_j)|\},
\end{gather*}
essentially approximating the analytic Riemann integral with its associated right Riemann sum.

We note that, computationally (as implemented in the paper's codebase), the $p_T^{\text{upd}}(\vy^j|\vx)$ and $p_T^{\text{lag}}(\vy^j|\vx)$ PDFs in \eqref{eq:updatedlaggeddistr_OIR} are obtained by using the so-called adaptive-lag variant of the CGNS online smoother in Theorem~\ref{thm:onlinesmoother} \cite{andreou2026adaptive}. Essentially, at each new observation $\vx^N$, only the most influential or impactful updates, in terms of the information gain about the state estimation of $\vy^j$, are carried out in the recursive equations in \eqref{eq:onlinerecursive} when transitioning and assimilating the new observation $\vx^N$ onto ${\vx(s\leq N-1)}$. That is, using the information-theoretic criteria defined in \cite{andreou2026adaptive} to determine this adaptive-lag $L_N$ at $t_N=T$, we only carry out the updates in \eqref{eq:onlinerecursive} for $N-L_N\leq j\leq N$, while for $j\in\llbracket N-L_N-1\rrbracket$ we set $\vm{\ns}^{j,N}=\vm{\ns}^{j,N-1}$ and $\mr{\ns}^{j,N}=\mr{\ns}^{j,N-1}$.

\section{Proof of Theorem~\ref{thm:obj_backward_CIR_linear}} \label{sec:Back_CIR_Linear_Thm_Proof}

Here we provide the proof of Theorem~\ref{thm:obj_backward_CIR_linear}, utilizing the results from Appendix~\ref{sec:Tool_Backward_CIR}.

\begin{proof}
Under the conditionally linear structure with additive noise feedbacks in \eqref{eq:reduced_model} and assumptions which establish linear stochastic stability, the posterior distributions of filter and smoother enjoy a statistical attractor, where the filter and smoother equilibrium distributions are necessarily Gaussian \cite{chen2014information, sarkka2023bayesian, chen2015noisy, gardiner2009stochastic, sarkka2019applied}. The steady-state algebraic equations for the equilibrium filter (algebraic Riccati equation) and smoother (algebraic symmetric Sylvester equation) covariance matrices \cite{liptser2001statistics, kandil2003matrix}, denoted by $R_{\text{\nf}}^{\infty}$ and $R_{\text{\ns}}^{\infty}$ respectively, yield \cite{andreou2026adaptive, chen2014information, sarkka2023bayesian, chen2015noisy, gardiner2009stochastic, sarkka2019applied}:
\begin{align*}
    R_{\text{\nf}}^{\infty}&=\frac{\lambda^y(\sigma^x)^2+\sigma^x\sqrt{\Psi}}{(\lambda^x)^2},\\
    R_{\text{\ns}}^{\infty}&=\frac{(\sigma^y)^2}{2\big((\lambda^y)^2+(\sigma^y)^2/R_{\text{\nf}}^{\infty}\big)},
\end{align*}
where $\Psi:=(\lambda^y\sigma^x)^2+(\lambda^x\sigma^y)^2$. (For one-dimensional $x$ and $y$ these are scalar quantities.) As such, as $T$ grows unboundedly (and by extension $N$, the current number of observations for $x$, per our setup from the beginning of Appendix~\ref{sec:Online_Smoother}), we have that $R_{\text{\nf}}^{N}$, $R_{\text{\ns}}^{N-1,N}=O(1)$. This extends to the filter and smoother means, with $\mu_{\text{\nf}}^{N}$, $\mu_{\text{\ns}}^{N-1,N}=\Theta(1)$ if $f^x$ or $f^y$ are periodic and bounded, which does not alter the assertion of this theorem.

Note that the system in \eqref{eq:reduced_model} is a CGNS (see \eqref{eq:CGNS}), therefore the forward and (complete) backward variants of the CIR metric, \eqref{eq:forward_CIR_metric} and \eqref{eq:backward_CIR_metric}--\eqref{eq:complete_backward_CIR_metric} respectively, enjoy a Gaussian signal--dispersion decomposition \cite{andreou2026assimilative, cai2002mathematical}, as illustrated in Appendix~\ref{sec:Tool_Forward_CIR} and Appendix~\ref{sec:Tool_Backward_CIR}, respectively. Therefore, we can then make the following assertion for the complete backward one (recall \eqref{eq:lackinfoupdate_OIR}--\eqref{eq:approx_info_gains_cgns_OIR}):
\begin{align}
    \begin{split}
        \lim_{T'\to T^-}\delta(t_j,T')&\approx \mathcal{P}^{j}_T=\mathcal{P}\big(p_T^{\text{upd}}(\vy^j|\vx), p_T^{\text{lag}}(\vy^j|\vx)\big)\\
        &\approx \frac{1}{2}\big(\mu_{\text{\ns}}^{N-1,N}-\mu_{\text{\nf}}^{N-1}\big)^2\big(D^{j,N-2}\big)^2(R_{\text{\ns}}^{j,N-1})^{-1}+ \frac{1}{2}\big(Q^{j,N}-1-\log(Q^{j,N})\big),
    \end{split} \label{eq:approx_info_gain}
\end{align}
for each $j\in\llbracket N\rrbracket$, where for $T\gg 1$ (and $N\gg 1$) from \eqref{eq:covarianceratio_OIR} we have
\begin{equation*}
    Q^{j,N}\approx 1+\frac{R_{\text{\ns}}^{\infty}-R_{\text{\nf}}^{\infty}}{R_{\text{\ns}}^{j,\infty}}(D^{j,\infty})^2,
\end{equation*}
since in this scenario we have $R_{\text{\nf}}^{N-1}\to R_{\text{\nf}}^{\infty}$ and $R_{\text{\ns}}^{N-1,N}\to R_{\text{\ns}}^{\infty}$, which exist and are finite due to the existence of the filter and smoother posterior attractors, respectively. Therefore, it suffices to asymptotically analyze the update operators $D^{j,N-2}$ for $j\in\llbracket N\rrbracket$ as $T,N\gg 1$. For that, as stated at the beginning of Appendix~\ref{sec:Online_Smoother}, we assume $\dt$, defined either by the model simulation's numerical integration time step or the temporal rate at which observations of $x$ are obtained, is sufficiently small ($\dt\ll1$). For a continuously observed $x$, this is certainly true through the formal limit $\dt\to0^+$.

From Theorem~\ref{thm:onlinesmoother}, we observe the following for each $j\in\llbracket N\rrbracket$ when $T$, $N\gg 1$:
\begin{equation} \label{eq:update_matrix_expanded}
    D^{j,N-2}=\prod_{i=j}^{N-2} (1-G^{y,i}\dt)= O(1)(1-G^{y,\infty}\dt)^{N-j},
\end{equation}
where \cite{andreou2026adaptive}:
\begin{equation*}
    G^{y,\infty}=\frac{\sqrt{\Psi}}{(\lambda^x)^2R_{\text{\nf}}^{\infty}}(\sqrt{\Psi}+\lambda^y\sigma^x)>0.
\end{equation*}
Recalling that we assume $\dt\ll1$, Taylor-expansion-based manipulations reveal that for large $T$ and $N$ we have
\begin{equation*}
    (D^{j,N-2})^2=\Theta\big(e^{2G^{y,\infty}(j-N)\dt}\big)=\Theta\big(e^{2G^{y,\infty}(j\dt-T)}\big).
\end{equation*}
Furthermore, due to the existence of the filter and smoother equilibrium Gaussian statistics, it is immediate that
\begin{equation*}
    \big(O(1)=\!\big)\,\mu_{\text{\ns}}^{N-1,N}-\mu_{\text{\nf}}^{N-1}\to \mu_{\text{\ns}}^{\infty}-\mu_{\text{\nf}}^{\infty},
\end{equation*}
as $\mu_{\text{\ns}}^{\infty}-\mu_{\text{\nf}}^{\infty}$ exists and is finite by Theorem~\ref{thm:onlinesmoother}. As such, for each $j\in\llbracket N\rrbracket$ when $T$, $N\gg 1$, combining the previous results and specifically plugging \eqref{eq:update_matrix_expanded} in \eqref{eq:approx_info_gain} yields:
\begin{equation} \label{eq:approx_info_gain_asympt}
    \lim_{T'\to T^-}\delta(t_j,T')\approx \mathcal{P}^{j}_T=O(1)\left(\Theta\big(e^{2G^{y,\infty}(j\dt-T)}\big)R_{\text{\ns}}^{j,\infty}+Q^{j,N}-1-\log(Q^{j,N})\right),
\end{equation}
with
\begin{equation} \label{eq:covratio_asympt}
    Q^{j,N}=1+\frac{O(1)}{R_{\text{\ns}}^{j,\infty}}\Theta\big(e^{2G^{y,\infty}(j\dt-T)}\big).
\end{equation}
Since we consider $T\gg1$ (and so $N\gg1$), it is immediate from \eqref{eq:covratio_asympt} that the information gain in \eqref{eq:approx_info_gain_asympt} is predominantly signal-driven, i.e., the dispersion component is essentially negligible; a numerical validation of this statement is showcased in the case study in Section~\ref{sec:Climate_Tipping}. Therefore, it follows that,
\begin{equation} \label{eq:signaldriven}
    \mathcal{P}_T^j-\mathcal{P}^0_T\approx O(1)\Theta\big(e^{2G^{y,\infty}(j\dt-T)}\big)R_{\text{\ns}}^{j,\infty}.
\end{equation}
Recalling the assumption that \eqref{eq:reduced_model} enjoys linear stochastic stability, meaning its linear dynamics ensure geometric ergodicity \cite{mattingly2002ergodicity, chen2018rigorous}, the online smoother variance quickly relaxes toward the smoother equilibrium variance in $j$: $R_{\text{\ns}}^{j,\infty}\to R_{\text{\ns}}^{\infty}$. As a result, from this, \eqref{eq:signaldriven}, \eqref{eq:objective_OIR_length_cgns}, and
\begin{equation*}
    0<\frac{\displaystyle\int_0^T e^{r (t-T)}\rmd t}{\displaystyle \underset{t\in[0,T]}{\max}\big\{e^{r (t-T)}\big\}}=\frac{1}{r}-\frac{e^{-Tr}}{r},
\end{equation*}
for $r>0$ arbitrary, then for $T$ large enough we finally recover that
\begin{equation*}
   \frac{1}{r_1} \leq \uptau^{\,\text{\normalfont{b}}}_{\text{approx}}(T) \leq \frac{1}{r_2}, \quad T\gg 1,
\end{equation*}
for some $r_1$, $r_2>0$ which do not depend on $T$.
\end{proof}
}

\bibliographystyle{elsarticle-num-names} 
\bibliography{references}

\end{document}